\documentclass[10pt,twocolumn,letterpaper]{article}

\usepackage[pagenumbers]{cvpr} 

\definecolor{cvprblue}{rgb}{0.21,0.49,0.74}
\usepackage[pagebackref,breaklinks,colorlinks,allcolors=cvprblue]{hyperref}

\usepackage{graphicx}
\usepackage{booktabs}
\usepackage{multirow}
\usepackage{makecell}
\usepackage{multirow}
\usepackage{adjustbox}
\usepackage{colortbl}
\usepackage{amssymb}  
\usepackage{pifont}
\usepackage{fontawesome5}
\usepackage{afterpage}
\usepackage{mathrsfs}
\usepackage[hang,flushmargin]{footmisc}
\usepackage{textcomp}
\usepackage{tcolorbox} 

\tcbset{
    mybox/.style={
        colback=gray!10,        
        colframe=black,         
        width=\textwidth,       
        arc=5pt,                
        outer arc=5pt,
        boxrule=0.5pt,          
        left=5pt,               
        right=5pt,              
        top=5pt,                
        bottom=5pt              
    }
}

\newcommand{\Checkmark}{\ding{51}}  
\newcommand{\XSolidBrush}{\ding{55}}  

\def\paperID{823} 
\def\confName{CVPR}
\def\confYear{2025}

\newcommand\blfootnote[1]{%
    \begingroup 
    \renewcommand\thefootnote{}\footnote{#1}%
    \addtocounter{footnote}{-1}%
    \endgroup 
}

\title{Towards Universal Soccer Video Understanding}

\author{Jiayuan Rao$^{1,2*}$, Haoning Wu$^{1,2*}$, Hao Jiang$^{3}$, Ya Zhang$^{1}$, Yanfeng Wang$^{1\dagger}$, Weidi Xie$^{1\dagger}$\\[3pt]
  $^{1}$School of Artificial Intelligence, Shanghai Jiao Tong University, China \\[2pt]
  $^{2}$CMIC, Shanghai Jiao Tong University, China \hspace{0.3cm} $^{3}$Alibaba Group, China \\[3pt]
  \url{https://jyrao.github.io/UniSoccer/}
}

\begin{document}

\maketitle

\blfootnote{*: These authors contribute equally to this work. \\ 
    $\dagger$: Corresponding author.
}

\begin{abstract}

As a globally celebrated sport, soccer has attracted widespread interest from fans all over the world. 
This paper aims to develop a comprehensive multi-modal framework for soccer video understanding.
Specifically, we make the following contributions in this paper:
(i) we introduce \textbf{SoccerReplay-1988}, the largest multi-modal soccer dataset to date, featuring videos and detailed annotations from 1,988 complete matches, with an automated annotation pipeline;
(ii) we present an advanced soccer-specific visual encoder, \textbf{MatchVision}, which leverages spatiotemporal information across soccer videos and excels in various downstream tasks;
(iii) we conduct extensive experiments and ablation studies on event classification, commentary generation, and multi-view foul recognition.
MatchVision demonstrates state-of-the-art performance on all of them, substantially outperforming existing models, which highlights the superiority of our proposed data and model.
We believe that this work will offer a standard paradigm for sports understanding research. 
\end{abstract}

\vspace{-6pt}
\textit{``Football is one of the world's best means of communication. It is impartial, apolitical, and universal."}

\quad \quad \quad \quad \quad \quad \quad \textemdash\textemdash \textit{ Franz Beckenbauer (1945 - 2024)}
\vspace{-6pt}

\section{Introduction}
\label{sec:intro}

\begin{figure}[ht]
    \centering
    \includegraphics[width=\linewidth]{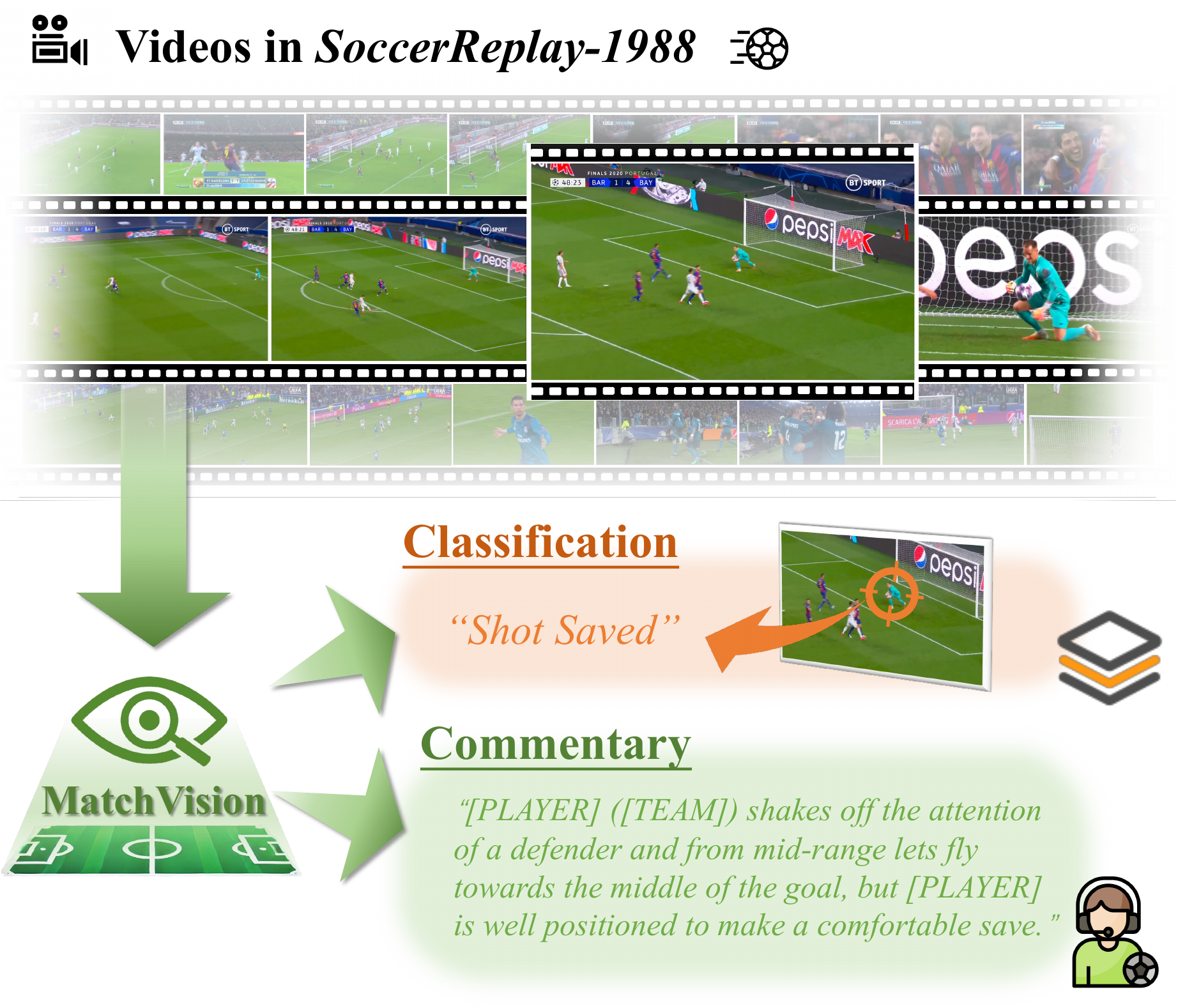}
    \vspace{-18pt}
    \caption{\textbf{Overview}. 
    We present \textbf{SoccerReplay-1988}, the largest soccer dataset to date, and a powerful soccer-specific visual encoder, \textbf{MatchVision}, capable of excelling in various tasks such as event classification and commentary generation.
    }
    \label{fig:teaser}
    \vspace{-12pt}
\end{figure}

Soccer, celebrated worldwide for its significant commercial value, has recently seen great research interest in integrating artificial intelligence~(AI) for soccer video understanding.
This is primarily motivated by the sport's complexity and the growing demand for enhanced analytics and improved viewing experiences. 
AI systems facilitate tactical analysis~\cite{wang2024tacticai}, allowing coaches to devise better strategies by uncovering patterns not apparent to the naked eye. In addition, it also supports automated content generation and enriches fan engagement through interactive and personalized content~\cite{goal, densecap, rao2024matchtimeautomaticsoccergame}. 
These capabilities promote a deeper understanding of soccer, simplify content creation, and foster a more engaging experience for fans and professionals.

Existing research in soccer video analysis primarily revolves around the SoccerNet series datasets~\cite{giancola2018soccernet1, deliege2021soccernet2, cioppa2022soccernet3}, which comprise 500 full-match videos for benchmarking various tasks, such as action spotting~\cite{giancola2018soccernet1, deliege2021soccernet2} and commentary generation~\cite{densecap, goal, rao2024matchtimeautomaticsoccergame}.
Despite this extensive coverage, the focus has predominantly been on designing specialized models for task-specific applications, leading to fragmented and incompatible solutions.
Such fragmentation underscores the need for a unified framework capable of integrating diverse demands, enabling more holistic and scalable advancements in soccer video understanding.

In this paper, we introduce \textbf{SoccerReplay-1988}, the largest and most comprehensive multi-modal soccer video dataset to date, featuring 1,988 complete match videos with rich annotations, such as event labels and textual commentaries. 
This dataset offers a solid foundation for developing advanced soccer understanding models and establishes a challenging new benchmark for the field.
Additionally, we have harmonized existing datasets to be compatible with ours, further expanding the available data resources.

Leveraging this dataset, we develop \textbf{MatchVision}, an advanced soccer-specific visual encoder tailored for diverse soccer understanding tasks. 
It employs the cutting-edge visual-language foundation model as the backbone, {\em e.g.}, SigLIP~\cite{zhai2023SigLIP}.
We further extend framewise visual features into spatiotemporal representations with temporal attentions~\cite{bertasius2021TimeSformer}, by training on diverse visual-language tasks on SoccerReplay-1988, as depicted in Figure~\ref{fig:teaser}. 
As a result, MatchVision exhibits strong adaptability across various tasks, such as event classification and commentary generation, serving as a universal and unified framework for comprehensive soccer video understanding.

To summarize, we make the following contributions in this paper: 
(i) we construct \textbf{SoccerReplay-1988}, the largest and most diverse soccer video dataset to date, featuring videos of 1,988 soccer matches with rich annotations, supported by an automated curation pipeline. 
This provides a solid foundation for developing robust and comprehensive soccer understanding models; 
(ii) we present {a powerful soccer-specific visual encoder, termed \textbf{MatchVision}, which effectively leverages spatiotemporal information in soccer videos, and can adapt to various tasks such as event classification and commentary generation, serving as a unified framework for soccer understanding;
(iii) we establish more comprehensive and challenging benchmarks based on our dataset, enabling more professional evaluation of soccer understanding models; 
(iv) extensive experiments and ablation studies demonstrate the superiority of our data and model across various downstream tasks, achieving state-of-the-art performance on both existing benchmarks and our newly established ones. 
We believe this work offers a viable paradigm for future sports video understanding.

\section{Related Works}
\label{sec:related_work}

\noindent \textbf{Sports Understanding}~\cite{thomas2017computervisionsports} is an evolving field that encompasses multiple research topics and integrates diverse data modalities, covering various tasks such as action spotting~\cite{giancola2018soccernet1, deliege2021soccernet2, gu2020fine}, commentary generation~\cite{densecap, yu2018fine, xi2025simple, xi2025eika, qi2019sports, rao2024matchtimeautomaticsoccergame}, athlete analysis~\cite{shao2020finegym, xu2022finediving}, tactical planning~\cite{wang2024tacticai}, sports health~\cite{ramkumar2022sports}, and intelligent refereeing~\cite{held2023vars, held2024XVARS}.
Furthermore, with the rapid development of multimodal large language models~(MLLMs), recent efforts~\cite{wu2022sportsvideoanalysislargescale, li2024sports, xia2024sportqa, xia2024sportu} have attempted to build more generalized frameworks to uniformly handle a variety of sports understanding tasks.

\vspace{2pt}
\noindent \textbf{Visual-Language Models}~\cite{CLIP, zhai2023SigLIP, li2022blip, li2023blip2, alayrac2022flamingo} have exhibited remarkable performance across extensive applications like classification, segmentation, image-text retrieval, and image captioning.
Recent efforts have ventured into more challenging video understanding~\cite{li2023llama-vid, li2023videochat2, zhang2023video-llama, zhou2025mlvubenchmarkingmultitasklong, shu2024videoxlextralongvisionlanguage} tasks, such as temporal alignment~\cite{li2024multi, han2022temporal}, dense captioning~\cite{zhou2024streaming, chen2024panda70m, yang2023vid2seq}, and audio description~\cite{han2023autoad, han2023autoad2, han2024autoad3}.
However, these efforts typically focus on general scenarios, limiting their adaptability to specific professional fields.
Thus, this paper aims to bridge this gap by advancing visual-language models tailored for comprehensive soccer understanding.

\vspace{2pt}
\noindent \textbf{Soccer Game Analysis}~\cite{cioppa2024soccernet2024challengesresults} has primarily focused on tasks such as action spotting~\cite{giancola2018soccernet1}, replay grounding~\cite{deliege2021soccernet2, baidu}, commentary generation~\cite{densecap, goal, rao2024matchtimeautomaticsoccergame}, player tracking~\cite{cioppa2022soccernet-track}, state reconstruction~\cite{somers2024soccernetgamestatereconstruction}, camera calibration~\cite{Cioppa_2021_CVPR_soccernet-camera, deliege2021soccernet2, cioppa2022soccernet-track} and foul recognition~\cite{held2023vars, held2024XVARS}, as facilitated by the SoccerNet~\cite{giancola2018soccernet1, deliege2021soccernet2, cioppa2022soccernet3, gautam2024soccernet-echoes} series datasets, with 500 full-match videos from 2015 to 2017.
Unlike existing methods that target designing specific models for distinct tasks, this paper aims to design a unified multi-modal framework that leverages spatiotemporal information within videos, serving as a specialized visual encoder for soccer video understanding.
\section{SoccerReplay-1988 Dataset}
To establish a solid foundation for soccer understanding, we construct \textbf{SoccerReplay-1988}, the largest soccer dataset to date.
Here, we first outline our data collection details and an overview of the dataset in Sec.~\ref{sec:data_collection}; 
followed by elaborating on our automated data curation pipeline in Sec.~\ref{sec:data_curation}; 
lastly, in Sec.~\ref{sec:statistics_discussion}, we present the data statistics and discussion.

\subsection{Dataset Collection}
\label{sec:data_collection}
To construct the \textbf{SoccerReplay-1988} dataset, we have collected untrimmed, full-match videos from the Internet, encompassing a total of 1,988 matches from six European major soccer leagues and championships\footnote{Premier (England), Laliga (Spain), Bundesliga (Germany), Serie-a (Italy), League-1 (France) and UEFA Champions League.}, spanning the {\em 2014-15} to {\em 2023-24} seasons.
For each match, we acquire textual commentaries with second-level timestamps from a sports text live website\footnote{flashscore.com}, with part of them annotated with specific event types such as \textit{corner} and \textit{goal}. Additionally, we also incorporate extensive metadata, including detailed background information about the games, players, coaches, referees, and teams, providing a solid foundation for future soccer understanding research.

We partition the SoccerReplay-1988 dataset into train, validation, and test sets, containing 1,488, 250, and 250 full-match videos with diverse and comprehensive annotations, respectively. 
These sets provide rich training data for downstream tasks, such as event classification and commentary generation, while establishing comprehensive and challenging benchmarks for soccer understanding, as further discussed in subsequent sections.

\begin{figure}[t]
    \centering
    \includegraphics[width=\linewidth]{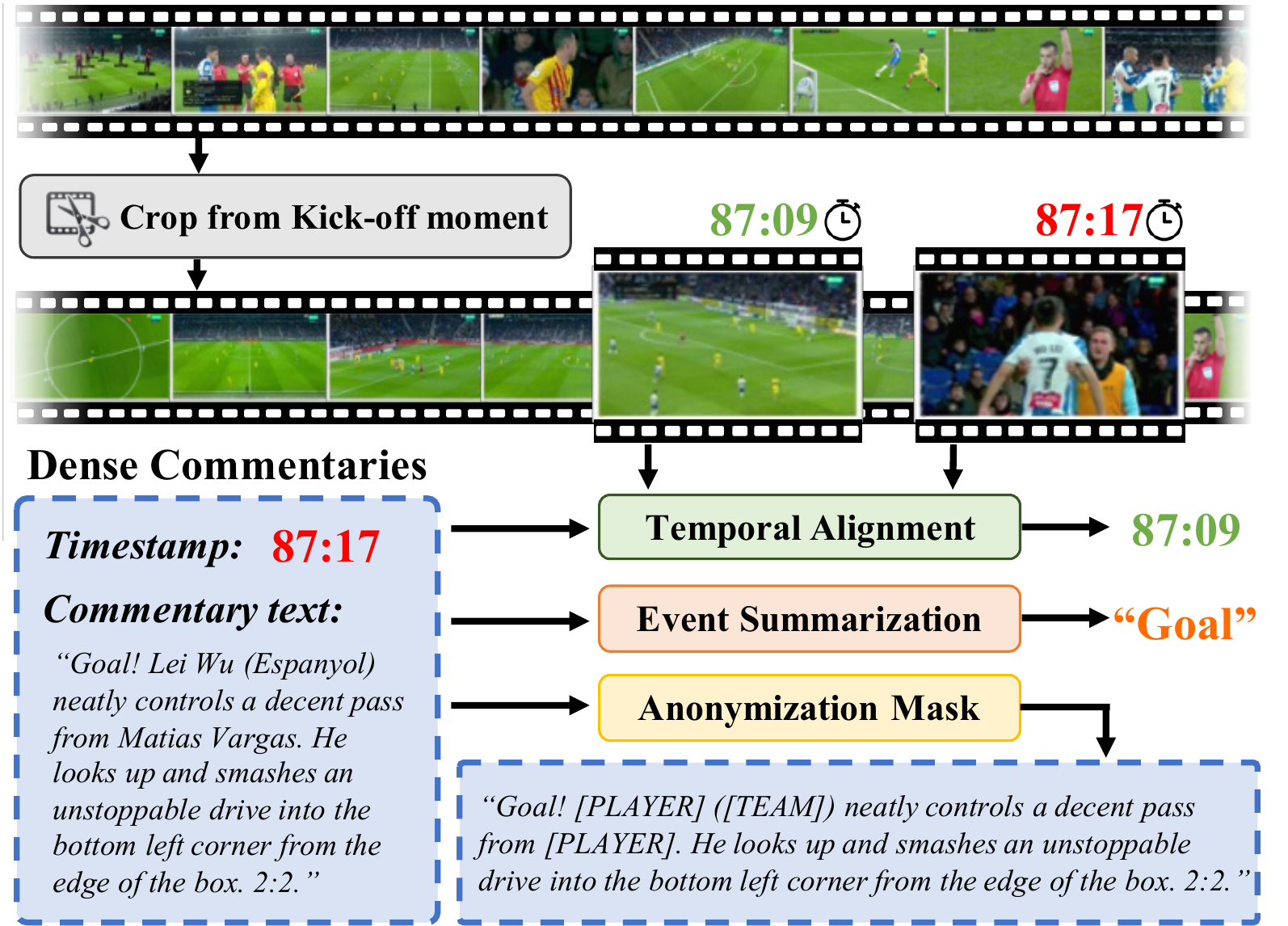}
    \vspace{-12pt}
    \caption{
    \textbf{Automated Data Curation Pipeline.} 
    The collected soccer video data are automatically processed for temporal alignment, event summarization, and anonymization by our curation pipeline.
    }
    \label{fig:curation}
    \vspace{-12pt}
\end{figure}

\subsection{Automated Data Curation}
\label{sec:data_curation}
Given the potential noise in raw data, such as irrelevant video content, inaccurate timestamps, and incomplete event annotations, we design an automated data curation pipeline, comprising (i) temporal alignment, (ii) event summarization, and (iii) anonymization, as illustrated in Figure~\ref{fig:curation}.

\vspace{2pt}
\noindent \textbf{Temporal Alignment.}
Here, we divide match videos into two halves, each starting at kick-off, and adopt the temporal alignment model from MatchTime~\cite{rao2024matchtimeautomaticsoccergame}, to synchronize textual commentary timestamps with those of video frames.

\vspace{2pt}
\noindent \textbf{Event Summarization.}
For samples without event annotations, we leverage LLaMA-3-70B~\cite{llama3} to summarize the events based on textual commentaries.
Concretely, we have expanded the event categories from 17 in SoccerNet~\cite{deliege2021soccernet2} to 24 types, 
for finer-grained soccer understanding, 
for example, categorizing penalties into {\em scored} and {\em missed}, and integrating modern soccer regulations like VAR. 
The resulting 24 event labels include: {\em `corner'}, {\em `goal'}, {\em `injury'}, {\em `own goal'}, {\em `penalty'}, {\em `penalty missed'}, {\em `red card'}, {\em `second yellow card'}, {\em `substitution'}, {\em `start of game~(half)'}, {\em `end of game~(half)'}, {\em `yellow card'}, {\em `throw in'}, {\em `free kick'}, {\em `saved by goal-keeper'}, {\em `shot off target'}, {\em `clearance'}, {\em `lead to corner'}, {\em `off-side'}, {\em `var'}, {\em `foul (no card)'}, {\em `statistics and summary'}, {\em `ball possession'}, and {\em `ball out of play'}.
More details on the used prompts are provided in the \textbf{Appendix}.

\vspace{2pt}
\noindent \textbf{Anonymization.}
Similar to~\cite{densecap}, we extract all person and team entity names from the metadata of \textbf{SoccerReplay-1988}, and replace them in textual commentaries with placeholders, such as ``[PLAYER]'', ``[TEAM]'', ``[COACH]'', and ``[REFEREE]'', ensuring consistency across tasks.

Moreover, our data curation pipeline can seamlessly extend to existing datasets, converting the SoccerNet series~\cite{deliege2021soccernet2, densecap} into our unified data format, termed {\em SoccerNet-pro}. 
This expansion further enlarges the standardized datasets available for soccer understanding tasks.

\begin{table}[t]
    \centering
    \begin{adjustbox}{max width=\linewidth}
    \small  
    \setlength{\tabcolsep}{0.10cm} 
    \begin{tabular}{lccccc}
    \toprule
    \cellcolor{white}\textbf{} & \multicolumn{5}{c}{\textbf{Existing Datasets}} \\
    \cmidrule(lr){2-6}
    \textbf{} & \textbf{\# Game} & \textbf{Duration(h)} & \textbf{\# Event} & \textbf{\# Anno.} & \textbf{\# Com.}\\
    \midrule
    SoccerNet-v1~\cite{giancola2018soccernet1} & 500 & 764 & 7 & 6.7k & - \\
    SoccerNet-v2~\cite{deliege2021soccernet2} & 500 & 764 & 17 & 110k  & - \\
    MatchTime~\cite{rao2024matchtimeautomaticsoccergame} & 471 & 716 & 14 & 14k  & 37k \\
    GOAL~\cite{goal} & 20 & 25.5 & - & - & 8.9k \\
    \midrule
    \cellcolor{white}\textbf{} & \multicolumn{5}{c}{\textbf{Our Curated Datasets}} \\
    \cmidrule(lr){2-6}
    \textbf{} & \textbf{\# Game} & \textbf{Duration(h)} & \textbf{\# Event} & \textbf{\# Anno.} & \textbf{\# Com.} \\
    \midrule
    SoccerNet-pro & 500 & 764 & 24 & 102k & 37k \\
    \textbf{SoccerReplay-1988} & \textbf{1,988} & \textbf{3,323} & \textbf{24} & \textbf{150k} & \textbf{150k} \\
    \midrule
    \rowcolor[gray]{0.9} \textbf{Integrated} & \textbf{2,488} & \textbf{4,087} & \textbf{24} & \textbf{252k}  & \textbf{187k} \\
    \bottomrule
    \end{tabular}
    \end{adjustbox}
    \vspace{-3pt}
    \caption{
    \textbf{Statistics of Soccer Datasets.} 
    Our SoccerReplay-1988 significantly surpasses existing datasets in both scale and diversity. 
    Here, \# Anno. and \# Com. refer to the number of event annotations and textual commentaries, respectively.
    }
    \label{tab:dataset_statistics}
    \vspace{-12pt}
\end{table}

\subsection{Statistics \& Discussion}
\label{sec:statistics_discussion}

\noindent\textbf{Dataset Statistics.}
As shown in Table~\ref{tab:dataset_statistics}, our dataset encompasses 3,323 hours of footage from 1,988 soccer matches, with an average duration of 100.3 minutes per match. 
The videos range in resolution from 360p to 720p and frame rates between 25 and 30 FPS. 

For textual annotations, this dataset features approximately 150K commentaries, averaging 76 per match, precisely temporal-aligned by the robust alignment model from MatchTime~\cite{rao2024matchtimeautomaticsoccergame}.
These commentaries cover 4,467 unique words, significantly surpassing the 2,873 words in existing datasets~\cite{densecap, rao2024matchtimeautomaticsoccergame}, greatly enriching textual diversity.
Automated event summarization based on these commentaries has yielded about 150K event annotations. 
Notably, a random sampling of 2\% of the data yields 98\% manual verification accuracy, ensuring high-quality automated labeling.

\vspace{2pt}
\noindent\textbf{SoccerReplay-test Benchmark.}
To facilitate a more comprehensive evaluation of soccer understanding models, we integrate 250 matches from SoccerReplay-1988 with 50 matches from the curated SoccerNet-pro, establishing \textbf{SoccerReplay-test}, a more challenging benchmark for event classification and commentary generation. 
This benchmark features nearly four times larger than existing datasets and comprises finer-grained event labels, richer textual commentaries, and up-to-date soccer regulations.

\begin{figure*}[t]
    \centering
    \includegraphics[width=\textwidth]{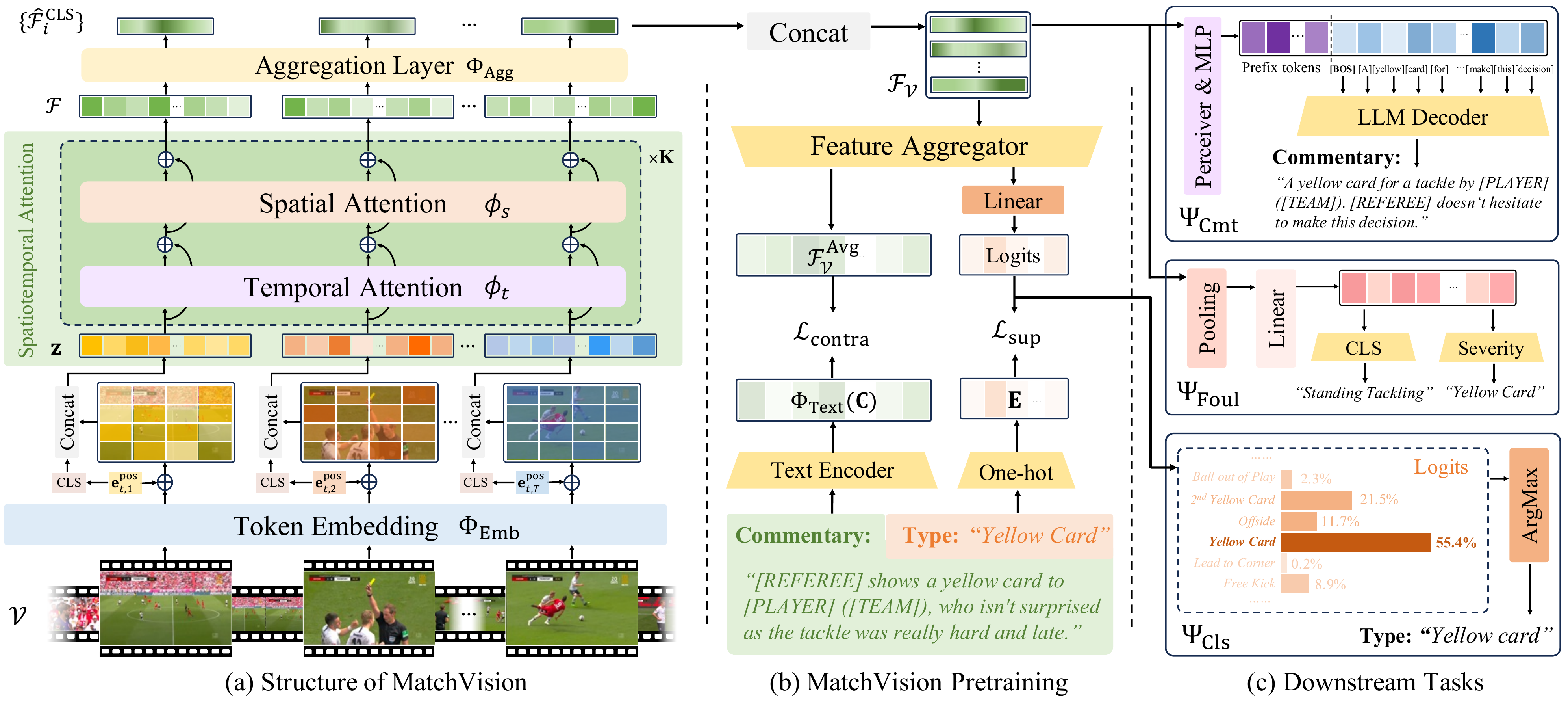}
    \vspace{-18pt}
    \caption{\textbf{Overview of MatchVision.} 
    (a) The model architecture and its spatiotemporal feature extraction process;
    (b) Details of visual encoder pretraining, including supervised classification and video-language contrastive learning;
    (c) Implementation details of specific heads for various downstream tasks, including commentary generation, foul recognition, and event classification.
    }
    \label{fig:structure}
    \vspace{-9pt}
\end{figure*}

\vspace{2pt}
\noindent\textbf{Discussion.}
To summarize, \textbf{SoccerReplay-1988} exhibits advancements in three aspects:
(i) it is the largest soccer video dataset to date, with nearly four times more videos than existing datasets;
(ii) it features more professional and diverse annotations, more suitable for fine-grained and comprehensive soccer understanding tasks;
(iii) It employs an automated curation pipeline for annotations and is thus scalable to provide a solid data foundation for future research.
All data from SoccerReplay-1988, including videos and annotations, are open-source for \textbf{non-commercial use}.
\section{Method}
\label{sec:method}

In this paper, we aim to develop a soccer-specific visual encoder, \textbf{MatchVision}, tailored for diverse soccer video analysis tasks.
We start by outlining our problem formulation in Sec.~\ref{sec:problem_formulation}. 
Next, in Sec.~\ref{sec:architecture}, we detail the architecture of MatchVision. 
The training procedures are thoroughly discussed in Sec.~\ref{sec:pretraining}. 
Finally, we describe the configurations for downstream tasks in Sec.~\ref{sec:downstream}, demonstrating the practical applications and effectiveness of our model.

\subsection{Problem Formulation}
\label{sec:problem_formulation}

In this work, we tackle the challenge of analyzing soccer video segments, denoted as $\mathcal{V} \in \mathbb{R}^{T \times 3 \times H \times W}$. 
Our goal is to utilize the visual encoder~($\Phi_{\mathrm{MatchVision}}$) to extract spatiotemporal features from these segments, which are then processed by multiple task-specific heads, formulated as:
\begin{align*}
    {\mathbf{E}, \mathbf{C}, \mathbf{F}} = \Psi(\Phi_{\mathrm{MatchVision}}(\mathcal{V}))
\end{align*}
Here, $\Psi = \{ \Psi_{\mathrm{cls}}, \Psi_{\mathrm{Cmt}}, \Psi_{\mathrm{Foul}}\}$ represents the task-specific heads, with $\mathbf{E}$, $\mathbf{C}$, and $\mathbf{F}$ denoting the output event types, textual commentaries, and foul types, respectively. 
This unified framework effectively learns relevant spatiotemporal features, and enables seamless integration across various downstream tasks for comprehensive soccer understanding.

\subsection{Architecture}
\label{sec:architecture}
\textbf{MatchVision} comprises three key components: 
(i) Token Embedding, 
(ii) Spatiotemporal Attention Block, and 
(iii) Aggregation Layer, as depicted in Figure~\ref{fig:structure}.

\vspace{2pt} 
\noindent \textbf{Token Embedding.}
In accordance with the convention in Vision Transformer~\cite{dosovitskiy2021ViT}, 
each frame~($\mathcal{I}_i$) from the video segment~($\mathcal{V} = \{ \mathcal{I}_1, \mathcal{I}_2, \cdots, \mathcal{I}_T \}$) is divided into $M$ non-overlapping patches of size $P\times P$ that span the entire frame. 

These patches are flattened into vectors~($\mathbf{x}_i^{p}$), 
where $p$ and $i$ denote the spatial and temporal positions, respectively.
Each vector is transformed via an embedding layer~($\Phi_\mathrm{Emb}$) into a token vector of size $\mathbb{R}^{1 \times D}$, and then added with a spatial position embedding~($\mathbf{e}_\mathrm{s}^{\mathrm{pos}} \in \mathbb{R}^{M \times D}$).
Subsequently, we concatenate a \texttt{[cls]} token along with each frame.
Finally, a temporal positional embedding~($\mathbf{e}_\mathrm{t}^{\mathrm{pos}} \in \mathbb{R}^{ T \times D}$) is added across features of all frames, 
as formulated below:
\begin{align*}
    & \mathbf{y}_i = [\mathbf{x}_{i}^{\mathrm{cls}}, \text{ } \Phi_{\mathrm{Emb}}([\mathbf{x}_{i}^{1}, \cdots, \mathbf{x}_{i}^{M}]) + \mathbf{e}_\mathrm{s}^{\mathrm{pos}}] \\
    & \mathbf{z} = [\mathbf{y}_1, \cdots, \mathbf{y}_T] + \mathbf{e}_\mathrm{t}^{\mathrm{pos}}
\end{align*}
Here, [$\cdot , \cdot$] denotes concatenation, and
$\mathbf{y}_i \in \mathbb{R}^{(M+1) \times D}$ represents the frame-wise features. 
The embedded features~($\mathbf{z}$) will then serve as input for spatiotemporal attention blocks.

\vspace{2pt} 
\noindent \textbf{Spatiotemporal Attention Block.}
Similar to TimeSformer~\cite{bertasius2021TimeSformer}, we utilize interleaved temporal and spatial attention to integrate spatiotemporal information in soccer videos. 
Concretely, each spatiotemporal attention block comprises a temporal self-attention layer and a spatial self-attention layer, 
{\em i.e.}, $\phi_{t}(\cdot)$ and $\phi_{s}(\cdot)$, respectively.

Given a video feature~($\mathbf{z} \in \mathbb{R}^{T \times (M+1) \times D}$),
we alternate temporal and spatial attention:
temporal attention facilitates interactions among tokens at the same spatial positions across distinct frames, while spatial attention enables interactions among tokens within the same frame.
Residual connections are employed in each layer.
After passing through a total of $\mathbf{K}$ spatiotemporal attention blocks, the resulting feature~($\mathcal{F}$) captures both intra-frame and inter-frame relationships, {\em i.e.}, $\mathcal{F} = [\phi_{s}(\phi_t(\mathbf{z}))]^{\mathbf{K}} \in \mathbb{R}^{ T\times(M+1) \times D}$.

\vspace{2pt}
\noindent \textbf{Aggregation Layer.}
To obtain video-level features, we employ an aggregation layer on the frame-wise spatiotemporal features. 
Specifically, for the $i$-th frame, we utilize spatial self-attention to aggregate information into its \texttt{[cls]} token, denoted as $\hat{\mathcal{F}}_i^{\mathrm{cls}} = \Phi_{\mathrm{Agg}}(\mathcal{F}_i)$. 
Concatenating the \texttt{[cls]} tokens of all frames yields the final video feature~($\mathcal{F}_{\mathcal{V}}$), that effectively encapsulates spatiotemporal characteristics of soccer video segments, thus enabling it to be applicable for various downstream soccer understanding tasks. 
This process can be formulated as:
\begin{align*}
    \mathcal{F}_{\mathcal{V}} = \Phi_{\mathrm{MatchVision}}(\mathcal{V}) = [\hat{\mathcal{F}}_1^{\mathrm{cls}}, \cdots, \hat{\mathcal{F}}_T^{\mathrm{cls}}] \in \mathbb{R}^{T \times D}
\end{align*}

\subsection{MatchVision Pretraining}
\label{sec:pretraining}

In this part, we aim to pretrain the visual encoder with triplet samples~($\{\mathcal{V}, \mathbf{E}, \mathbf{C}\}$), comprising videos, event labels, and textual commentaries. 
Concretely, we investigate two distinct pretraining strategies: supervised classification and video-language contrastive learning.

\vspace{2pt}
\noindent \textbf{Supervised Classification.}
One way to pretrain the visual encoder is supervised learning on event classification.
To be specific, the extracted visual features~($\mathcal{F}_{\mathcal{V}}$) are aggregated by a temporal self-attention layer into a learnable \texttt{[cls]} token, denoted as $\mathcal{F}_{\mathcal{V}}^{\mathrm{cls}}$, similar to the spatial-wise aggregation mentioned above.
This token is then fed into a linear classifier, and trained with a cross-entropy loss for event classification. 
The objective is denoted as $\mathcal{L}_{\mathrm{sup}}$.

\vspace{2pt}
\noindent \textbf{Video-Language Contrastive Learning.}
As an alternative, we can also pretrain our visual encoder with video-text contrastive learning. 
Specifically, we adopt simple average pooling on the video feature to get the aggregated visual feature~($\mathcal{F}_{\mathcal{V}}^{\mathrm{Avg}}$), and encode the textual commentary~($\mathbf{C}$) with a text encoder~($\Phi_{\mathrm{Text}}$).
We train the model with sigmoid loss~($\mathcal{L}_{\mathrm{sigmoid}}$), as used in SigLIP~\cite{zhai2023SigLIP}.
Note that, some video clips may have highly similar commentaries, 
for example, {\em `start of the game'}, 
we treat the commentaries with high similarity in the same batch as positive samples when calculating loss functions. This can be expressed as follows:
\begin{align*}
    \mathcal{L}_{\mathrm{contra}} = \mathcal{L}_{\mathrm{sigmoid}}(\mathcal{F}_{\mathcal{V}}^{\mathrm{Avg}}, \Phi_{\mathrm{Text}}(\mathbf{C}))
\end{align*}

\subsection{Downstream Tasks}
\label{sec:downstream}
After the pretraining mentioned above, MatchVision can now serve as a versatile visual encoder, to map the soccer video segments into visual features~($\mathcal{F}_{\mathcal{V}}$), for training task-specific heads $\Psi = \{ \Psi_{\mathrm{cls}}, \Psi_{\mathrm{Cmt}}, \Psi_{\mathrm{Foul}} \}$ across different downstream tasks, 
including: (i) event classification, (ii) commentary generation, and (iii) foul recognition.

\vspace{2pt} 
\noindent \textbf{Event Classification.}
Similar to supervised classification above, we concatenate a learnable \texttt{[cls]} token to aggregate frame-wise visual features via temporal self-attention. 
This token is then fed into a linear classifier for event classification.
The event classification head~($\Psi_{\mathrm{cls}}$) is trained with a cross-entropy loss while freezing the visual encoder.

\vspace{2pt} 
\noindent \textbf{Commentary Generation.} 
We follow the paradigm in MatchTime~\cite{rao2024matchtimeautomaticsoccergame} to generate anonymized textual commentary for soccer video clips.
Concretely, the commentary generation head~($\Psi_{\mathrm{Cmt}}$) employs a Perceiver~\cite{jaegle2021perceiver} aggregator to consolidate visual features, which are then projected by a trainable MLP, serving as prefix embeddings for a large language model~(LLM).
Subsequently, an off-the-shelf LLM decodes these embeddings into textual commentary. 
We adopt the negative log-likelihood loss, commonly used for auto-regressive next-token prediction.

\vspace{2pt}
\noindent \textbf{Foul Recognition.}
As outlined in~\cite{held2023vars}, the foul recognition task takes multi-view videos from the same scene as inputs, with each sample annotated with a foul class~(8 types) and severity~(4 levels).
We encode these multi-view videos with MatchVision, and aggregate the extracted features into a single feature vector, via either max or average pooling, following the common practice.
Subsequently, the foul recognition head~($\Psi_{\mathrm{Foul}}$) employs a shared MLP and two task-specific linear classifiers, to predict foul type and severity, respectively.
Similar to event classification, we use the combination of cross-entropy losses on the foul type and severity classification to jointly train $\Psi_{\mathrm{Foul}}$.

\vspace{2pt}
\noindent \textbf{Discussion.}
Pretraining MatchVision on large-scale soccer data equips it with substantial domain-specific knowledge, enabling it to serve as a universal visual encoder adaptable to various downstream soccer understanding tasks.
\begin{table*}[t]
    \centering
    \begin{adjustbox}{max width=\textwidth}
    \small  
    \begin{tabular}{lcccccccccccc}
    \toprule
    \multirow{2}{*}{\centering \makecell{\textbf{Visual Encoder}}} 
    & \multicolumn{3}{c}{\textbf{Dataset}} & \multicolumn{3}{c}{\textbf{Classification (\%)}} & \multicolumn{5}{c}{\textbf{Commentary}} \\
    \cmidrule(lr){2-4} \cmidrule(lr){5-7} \cmidrule(lr){8-12}
    & \textbf{SN} & \textbf{MT} & \textbf{SR} & \textbf{Acc.@1} & \textbf{Acc.@3} & \textbf{Acc.@5} & \textbf{B@1} & \textbf{B@4} & \textbf{M} & \textbf{R-L} & \textbf{C} \\
    \midrule
    \multicolumn{12}{c}{Off-the-shelf Models} \\
    \midrule
    I3D~\cite{i3d} & \XSolidBrush & \XSolidBrush & \XSolidBrush & 45.4 & 82.5 & 93.2 & 26.77 & 5.57 & 24.17 & 23.12 & 18.73 \\
    C3D~\cite{c3d} & \XSolidBrush & \XSolidBrush & \XSolidBrush & 47.8 & 85.1 & 95.0 & \textcolor{red}{\textbf{28.13}} & \textcolor{blue}{\underline{6.64}} & 24.52 & 24.23 & 27.88 \\
    ResNet~\cite{he2016resnet} & \XSolidBrush & \XSolidBrush & \XSolidBrush & 47.2 & 84.6 & 94.4 & 27.34 & 6.57 & 24.72 & 24.43 & 27.29 \\
    CLIP~\cite{CLIP} & \XSolidBrush & \XSolidBrush & \XSolidBrush & 48.5 & 85.5 & 95.2 & 26.25 & 6.51 & 24.27 & 24.75 & 28.17 \\
    InternVideo~\cite{wang2022internvideo} & \XSolidBrush & \XSolidBrush & \XSolidBrush & 49.9 & \textcolor{red}{\textbf{87.0}} & \textcolor{red}{\textbf{95.9}} & 27.12 & 6.54 & \textcolor{blue}{\underline{25.02}} & \textcolor{blue}{\underline{24.82}} & \textcolor{blue}{\underline{29.90}} \\
    SigLIP~\cite{zhai2023SigLIP} & \XSolidBrush & \XSolidBrush & \XSolidBrush & \textcolor{red}{\textbf{50.2}} & \textcolor{blue}{\underline{86.7}} & \textcolor{blue}{\underline{95.6}} & \textcolor{blue}{\underline{27.85}} & \textcolor{red}{\textbf{6.98}} & \textcolor{red}{\textbf{25.16}} & \textcolor{red}{\textbf{25.03}} & \textcolor{red}{\textbf{31.38}} \\
    \midrule
    \multicolumn{12}{c}{Pretrain with Supervised Classification} \\
    \midrule
    Baidu~\cite{baidu} & \Checkmark & \XSolidBrush & \XSolidBrush & 56.4 & 91.9 & 97.3 & \textcolor{red}{\textbf{31.20}} & \textcolor{blue}{\underline{8.88}} & \textcolor{blue}{\underline{26.56}} & \textcolor{blue}{\underline{26.61}} & \textcolor{blue}{\underline{38.93}} \\
    SigLIP & \Checkmark & \XSolidBrush & \XSolidBrush & 55.9 & 89.6 & 94.9 & 28.51 & 7.39 & 25.96 & 25.94 & 35.71 \\
    SigLIP & \Checkmark & \Checkmark & \Checkmark & 57.9 & 91.7 & 97.5 & 30.95 & 8.56 & 25.79 & 26.17 & 38.24 \\
    \textbf{MatchVision} & \Checkmark & \XSolidBrush & \XSolidBrush & \textcolor{blue}{\underline{82.5}} & \textcolor{blue}{\underline{96.6}} & \textcolor{blue}{\underline{98.8}} & 29.45 & 7.92 & 26.01 & 26.21 & 36.15 \\
    \textbf{MatchVision} & \Checkmark & \Checkmark & \Checkmark & \textcolor{red}{\textbf{84.0}} & \textcolor{red}{\textbf{97.3}} & \textcolor{red}{\textbf{99.2}} & \textcolor{blue}{\underline{31.05}} & \textcolor{red}{\textbf{9.06}} & \textcolor{red}{\textbf{26.94}} & \textcolor{red}{\textbf{27.93}} & \textcolor{red}{\textbf{42.20}} \\
    \midrule
    \multicolumn{12}{c}{Pretrain with Visual-Language Contrastive Learning} \\
    \midrule
    SigLIP & \XSolidBrush & \Checkmark & \XSolidBrush & 55.4 & 88.8 & 97.0 & 28.72 & 7.72 & 25.91 & 26.17 & 32.27 \\
    SigLIP & \XSolidBrush & \Checkmark & \Checkmark & \textcolor{blue}{\underline{66.8}} & \textcolor{blue}{\underline{93.7}} & \textcolor{blue}{\underline{98.6}} & \textcolor{blue}{\underline{30.35}} & \textcolor{blue}{\underline{8.12}} & \textcolor{blue}{\underline{26.05}} & \textcolor{blue}{\underline{26.38}} & \textcolor{blue}{\underline{39.41}} \\
    \textbf{MatchVision} & \XSolidBrush & \Checkmark & \XSolidBrush & 58.9 & 89.0 & 97.1 & 30.33 & 7.97 & 25.48 & 26.33 & 33.87 \\
    \textbf{MatchVision} & \XSolidBrush & \Checkmark & \Checkmark & \textcolor{red}{\textbf{67.9}} & \textcolor{red}{\textbf{93.9}} & \textcolor{red}{\textbf{98.6}} & \textcolor{red}{\textbf{31.94}} & \textcolor{red}{\textbf{9.12}} & \textcolor{red}{\textbf{26.24}} & \textcolor{red}{\textbf{27.56}} & \textcolor{red}{\textbf{40.76}} \\
    \midrule
    \multicolumn{12}{c}{Pretrain with Hybrid Supervised-Contrastive Training} \\
    \midrule
    SigLIP & \Checkmark & \Checkmark & \XSolidBrush & 71.2 & 94.5 & 98.7 & 28.63 & 7.82 & 25.74 & 25.35 & 34.09 \\
    SigLIP & \Checkmark & \Checkmark & \Checkmark & 67.1 & 93.2 & 98.1 & \textcolor{blue}{\underline{30.71}} & \textcolor{blue}{\underline{8.78}} & \textcolor{blue}{\underline{26.26}} & \textcolor{blue}{\underline{26.74}} & \textcolor{blue}{\underline{41.82}} \\
    \textbf{MatchVision} & \Checkmark & \Checkmark & \XSolidBrush & \textcolor{blue}{\underline{76.4}} & \textcolor{blue}{\underline{96.0}} & \textcolor{blue}{\underline{99.0}} & 30.65 & 8.33 & 25.28 & 26.31 & 37.23 \\
    \textbf{MatchVision} & \Checkmark & \Checkmark & \Checkmark & \textcolor{red}{\textbf{80.1}} & \textcolor{red}{\textbf{97.1}} & \textcolor{red}{\textbf{99.1}} & \textcolor{red}{\textbf{33.58}} & \textcolor{red}{\textbf{9.14}} & \textcolor{red}{\textbf{26.82}} & \textcolor{red}{\textbf{28.21}} & \textcolor{red}{\textbf{44.18}} \\
    \bottomrule
    \end{tabular}
    \end{adjustbox}
    \vspace{-3pt}
    \caption{
    \textbf{Quantitative Results on Event Classification and Commentary Generation.} 
    Here, SN, MT, and SR represent finetuning with curated SoccerNet-v2~\cite{deliege2021soccernet2}, MatchTime~\cite{rao2024matchtimeautomaticsoccergame}, and SoccerReplay-1988, respectively.
    B, M, R-L, and C refer to BLEU, METEOR, ROUGE-L, and CIDEr metrics, respectively.
    Within each unit, we denote the best performance in \textcolor{red}{\textbf{RED}} and the second-best performance in \textcolor{blue}{\underline{BLUE}}.
    }
    \label{tab:quantitative_results}
    \vspace{-12pt}
\end{table*}

\section{Experiments}
\label{sec:experiment}
This section begins with implementation details in Sec.~\ref{sec:implementation_details}; 
followed by quantitative evaluations across downstream tasks in Sec.~\ref{sec:quantitative_comparisons}; 
then, we conduct ablation studies on our SoccerReplay-test benchmark to analyze the effectiveness of the proposed dataset and model in Sec.~\ref{sec:ablation_studies};
finally, we provide qualitative results for comparison in Sec.~\ref{sec:qualitative_comparison}.

\subsection{Implementation Details}
\label{sec:implementation_details}
In our experiments, video segments are sampled at 1FPS around annotated timestamps, capturing a 30-second window for each sample. 
Frames are resized to $224 \times 224$ pixels as inputs.
We initialize the embedding layer, spatial attention layers, aggregation layer, and text encoder of MatchVision with pretrained weights from SigLIP~Base-16~\cite{zhai2023SigLIP} and adopt LLaMA-3~(8B)~\cite{llama3} as the off-the-shelf LLM decoder for commentary generation.
All experiments are conducted on $4\times$ Nvidia H800 GPUs with the AdamW~\cite{adamW} optimizer. 
Next, we elaborate on the training and evaluation details about visual encoder pretraining and downstream tasks.

\vspace{2pt}
\noindent \textbf{Visual Encoder Pretraining.}
For both pretraining strategies, we use a batch size of 40 for 15 epochs.
The learning rate for all randomly initialized modules, including the temporal attention layers, aggregator layer, and linear classifier, is set to $1 \times 10^{-4}$.
Meanwhile, the learning rate for modules initialized with pretrained parameters~(including the text encoder) is set to $5 \times 10^{-5}$.
In contrastive training, we adopt a multi-positive strategy where each textual commentary, based on its event label, considers closely related categories ({\em e.g. ``start of game''} and {\em ``offside''}) as positive samples.

\vspace{2pt}
\noindent \textbf{Downstream Tasks.}
In all downstream tasks, unless otherwise specified, we use the frozen visual encoder for feature extraction and only train the task-specific heads with a learning rate of $1 \times 10^{-4}$ for 30 epochs.
The batch sizes for event classification, commentary generation, and foul recognition are set to 40, 32, and 8, respectively.
We adopt specific evaluation metrics for these three tasks:
(i) For event classification, we use the top-1/3/5 classification accuracy; 
(ii) For commentary generation, we employ several commonly-used language evaluation metrics, including BLEU~\cite{papineni2002bleu}, METEOR~\cite{banerjee2005meteor}, ROUGE-L~\cite{lin2004rouge}, and CIDEr~\cite{vedantam2015cider};
(iii) For foul recognition, we follow the common practice, and report top-1/2 and top-1 accuracy for the foul type and severity classification, respectively.

\vspace{2pt}
\noindent \textbf{Benchmarks \& Baselines.}
To ensure fair and reliable comparisons with existing work, we evaluate event classification~(24 types) on 100 matches from curated SoccerNet-v2~\cite{deliege2021soccernet2} test set; 
commentary generation on 49 matches from SN-Caption-test-align benchmark manually aligned in~\cite{rao2024matchtimeautomaticsoccergame}; 
and foul recognition on MVFoul~\cite{held2023vars}.
We consider various baselines: for the first two tasks, this includes off-the-shelf general visual encoders such as ResNet~\cite{he2016resnet}, C3D~\cite{c3d}, I3D~\cite{i3d}, CLIP~\cite{CLIP}, SigLIP~\cite{zhai2023SigLIP}, and InternVideo~\cite{wang2022internvideo}, along with Baidu~\cite{baidu} and SigLIP finetuned with soccer-specific data.
For foul recognition, we follow previous work~\cite{held2023vars, held2024XVARS} and adopt ResNet~\cite{he2016resnet}, R(2+1)D~\cite{tran2018closer}, and MViT~\cite{fan2021mvit} jointly finetuned with classifiers, as baselines.

\begin{table}[t]
    \centering
    \small
    \renewcommand{\arraystretch}{1.0}
    \setlength{\tabcolsep}{4pt}
    \begin{tabular}{ccccccc}
        \toprule
        \multicolumn{3}{c}{\textbf{Visual Encoder}} & \multicolumn{2}{c}{\textbf{Foul Class}} & \multicolumn{2}{c}{\textbf{Severity}} \\
        \cmidrule(lr){1-3} \cmidrule(lr){4-5} \cmidrule(lr){6-7}
        \textbf{Backbone} & \textbf{Train} & \textbf{Agg.} & \textbf{Acc.@1} & \textbf{Acc.@2} & \textbf{Acc.@1} \\
        \midrule
        \multirow{2}{*}{ResNet~\cite{he2016resnet}} & \multirow{2}{*}{\ding{51}} & Mean & 0.31 & 0.56 & 0.34 \\
        & & Max & 0.32 & 0.60 & 0.32 \\
        \midrule
        \multirow{2}{*}{R(2+1)D~\cite{tran2018closer}} & \multirow{2}{*}{\ding{51}} & Mean & 0.31 & 0.55 & 0.34 \\
        & & Max & 0.32 & 0.56 & 0.39 \\
        \midrule
        \multirow{2}{*}{MViT~\cite{fan2021mvit}} & \multirow{2}{*}{\ding{51}} & Mean & 0.40 & 0.65 & 0.38 \\
        & & Max & \textbf{0.47} & \underline{0.69} & 0.43 \\
        \midrule
        \multirow{2}{*}{\textbf{MatchVision}} & \multirow{2}{*}{\ding{55}} & Mean & \underline{0.44} & 0.53 & \textbf{0.58} \\
        & & Max & 0.35 & \textbf{0.70} & \underline{0.46} \\
        \bottomrule
    \end{tabular}
    \vspace{-3pt}
    \caption{
    \textbf{Quantitative Results on Multi-view Foul Recognition.} 
    Our frozen MatchVision encoder can achieve comparable performance with other jointly finetuned visual encoders.
    }
    \vspace{-9pt}
    \label{tab:quantitative_foul}
\end{table}

\subsection{Quantitative Evaluation}
\label{sec:quantitative_comparisons}
As depicted in Table~\ref{tab:quantitative_results}, we draw two observations on event classification and commentary generation:
(i) visual encoders trained on soccer data substantially outperform off-the-shelf general encoders~(ResNet, C3D, I3D, CLIP, and InternVideo), underscoring the necessity of building specialized models for soccer understanding;
(ii) almost all visual encoders, across all training settings, benefit from \textbf{SoccerReplay-1988}, emphasizing the value of constructing large-scale, high-quality data for soccer understanding.
Next, we will delve into each task to discuss the results.

\vspace{2pt} 
\noindent \textbf{Event Classification.}
With identical training strategies and data, MatchVision considerably outperforms other methods in classification accuracy, demonstrating the superiority of its architecture, which effectively leverages spatiotemporal features within soccer videos. 
For example, MatchVision achieves a Top-1 accuracy of 82.5\%, significantly surpassing SigLIP's 55.9\% under the same training conditions.
Moreover, models trained via supervised classification excel others, primarily because the pre-training task shares the same objectives as the downstream event classification task.

\vspace{2pt} 
\noindent \textbf{Commentary Generation.}
Visual encoders trained with visual-language contrastive learning exhibit better commentary generation performance than those trained with supervised classification, as this strategy better captures correlations between visual and textual features. 
Additionally, while MatchVision trained solely on SoccerNet slightly underperforms Baidu~\cite{baidu}, incorporating SoccerReplay-1988 enables it to outperform on most metrics.
This demonstrates that MatchVision can take advantage of large-scale datasets.
Finally, a hybrid training approach, starting with supervised classification followed by visual-language contrastive learning, enables MatchVision to achieve optimal performance. 
This indicates that learning coarse-grained tasks such as classification provides a foundation for fine-grained tasks like commentary generation, and fully leveraging data unlocks the potential of soccer understanding.

\vspace{2pt}
\noindent \textbf{Foul Recognition.}
As demonstrated in Table~\ref{tab:quantitative_foul}, MatchVision achieves performance comparable to jointly finetuned state-of-the-art methods in foul recognition, even with a frozen visual encoder.
This highlights that MatchVision effectively learns substantial knowledge from large-scale soccer data and adapts seamlessly to downstream tasks.
Comparisons with additional baselines from the SoccerNet foul recognition challenges~\cite{cioppa2024soccernet2024challengesresults} are provided in the \textbf{Appendix}.

\begin{table}[t]
    \centering
    \small
    \renewcommand{\arraystretch}{1.0}
    \setlength{\tabcolsep}{4pt}
    \resizebox{.84\columnwidth}{!}{  
        \begin{tabular}{cccccc}
            \toprule
            \multicolumn{3}{c}{\textbf{Pretrain}} & \multicolumn{3}{c}{\textbf{Classification(\%)}} \\
            \cmidrule(lr){1-3} \cmidrule(lr){4-6}
            \textbf{Sup.} & \textbf{Contra.} & \textbf{SR} & \textbf{Acc.@1} & \textbf{Acc.@3} & \textbf{Acc.@5} \\
            \midrule
            \ding{51} & \ding{55} & \ding{55} & 62.67 & 83.00 & 89.81 \\
            \ding{51} & \ding{55} & \ding{51} & \textbf{68.03} & \textbf{86.90} & \textbf{92.38} \\
            \midrule
            \ding{55} & \ding{51} & \ding{55} & 46.97 & 75.53 & 85.85 \\
            \ding{55} & \ding{51} & \ding{51} & 57.41 & 83.13 & 91.00 \\
            \midrule
            \ding{51} & \ding{51} & \ding{55} & 56.86 & 80.30 & 88.09 \\
            \ding{51} & \ding{51} & \ding{51} & \underline{63.59} & \underline{85.21} & \underline{91.63} \\
            \bottomrule
        \end{tabular}
    }
    \vspace{-3pt}
    \caption{\textbf{Ablations on Event Classification.} 
    We explore the impact of various training settings of our MatchVision encoder on the SoccerReplay-test benchmark. 
    Here, Sup., Contra., and SR refer to supervised classification, visual-language contrastive learning, and the SoccerReplay-1988 dataset, respectively.
    }
    \vspace{-9pt}
    \label{tab:ablation_event_classification}
\end{table}

\begin{figure*}[t]
    \centering
    \includegraphics[width=\textwidth]{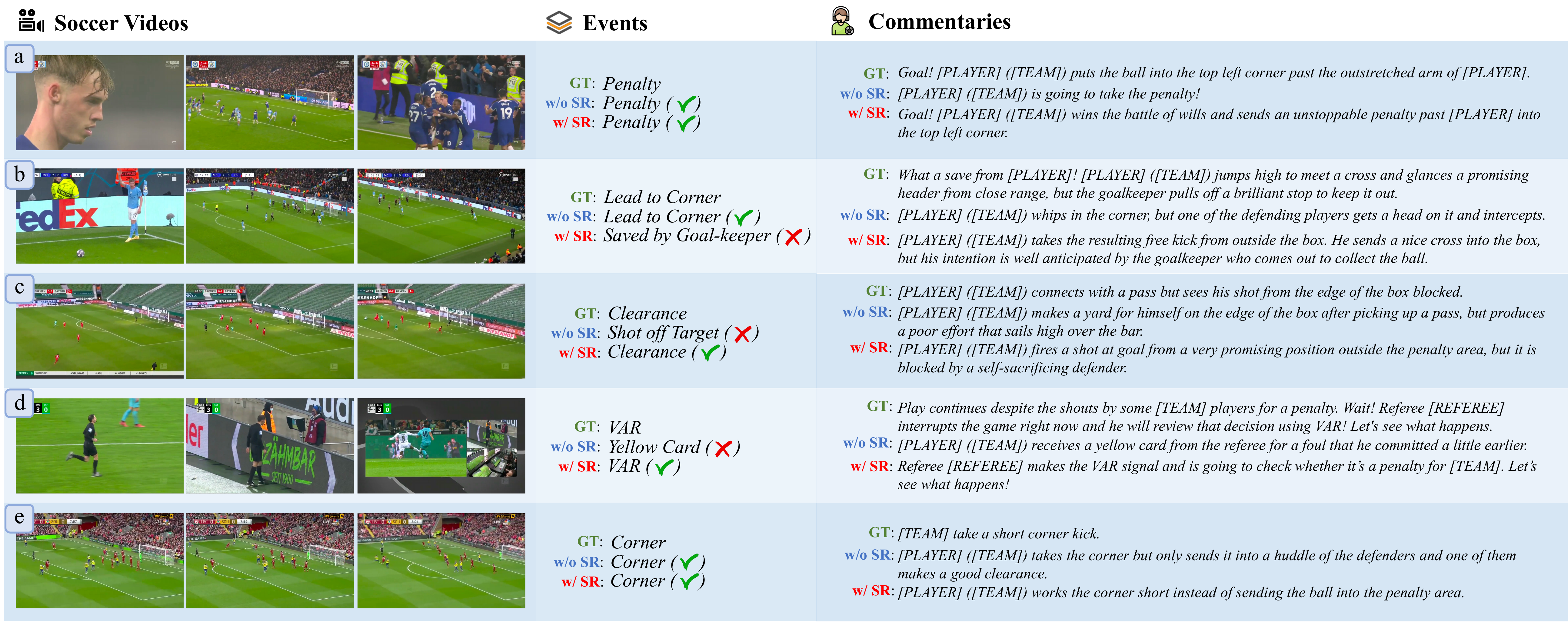}
    \vspace{-12pt}
    \caption{
    \textbf{Qualitative Results for Event Classification and Commentary Generation.} 
    Here, ``w/o SR'' and ``w/ SR'' indicate models trained without and with the SoccerReplay-1988 dataset, respectively. 
    Incorporating SoccerReplay-1988 improves event classification accuracy. 
    Moreover, this enriched training data enables the model to demonstrate several advantages in commentary generation: (a) more detailed descriptions, (b) greater linguistic variety, (c) higher event depiction accuracy, (d) better adherence to updated rules, and (e) improved specificity in scenario response.
    }
    \label{fig:qualitative}
    \vspace{-6pt}
\end{figure*}

\subsection{Ablation Studies}
\label{sec:ablation_studies}
We conduct ablation experiments on event classification and commentary generation using our \textbf{SoccerReplay-test} benchmark. 
These experiments validate the effectiveness of our proposed dataset and model, while establishing a baseline for future evaluations on this benchmark.

\vspace{2pt}
\noindent \textbf{Event Classification.}
We evaluate event classification on 300 matches from our \textbf{SoccerReplay-test} benchmark using the MatchVision visual encoder pretrained with various strategies.
Features are extracted by MatchVision and processed with a learnable aggregation layer and a linear classifier.
The default training set is our curated SoccerNet-pro.
As shown in Table~\ref{tab:ablation_event_classification}, integrating SoccerReplay-1988 for training results in significant performance improvements across all pretraining strategies, yielding the significance of our dataset.
Additionally, supervised classification outperforms visual-language contrastive learning and hybrid pretraining. 
This is due to its closer alignment with downstream event classification task, and the scale of event annotations is far larger than that of textual commentaries, further confirming the substantial benefits of data scaling for boosting soccer understanding.

\vspace{2pt}
\noindent \textbf{Commentary Generation.}
With the pretrained MatchVision encoder, we train the commentary generation head on the MatchTime~\cite{rao2024matchtimeautomaticsoccergame} and SoccerReplay-1988 datasets using various training strategies.
By default, only the Perceiver~\cite{jaegle2021perceiver} aggregation layer and projection layer within the head are trained.
For joint training with the LLM decoder, considering computational costs, we incorporate LoRA~\cite{hu2021lora} layers while freezing the original LLM layers.
As shown in Table~\ref{tab:ablation_commentary_generation}, incorporating SoccerReplay-1988 significantly improves performance on all metrics, confirming substantial advantages of our proposed dataset.
This performance gap also reflects the challenges of our established benchmark, which features diverse vocabulary, richer semantics, and updated soccer rules.
Additionally, jointly finetuning the visual encoder and the LLM decoder provides a feasible approach for further improvements.

\begin{table}[t]
    \centering
    \begin{adjustbox}{max width=\textwidth}
    \small  
    \begin{tabular}{ccccccc}
    \toprule
    \multicolumn{2}{c}{\textbf{Trainable}}& \multicolumn{5}{c}{\textbf{Commentary Metrics}} \\
    \cmidrule(lr){1-2} \cmidrule(lr){3-7}
    \textbf{V} & \textbf{L}& \textbf{B@1} & \textbf{B@4} & \textbf{M} & \textbf{R-L} & \textbf{C} \\
    \midrule
    \multicolumn{7}{c}{Trained on MatchTime} \\
    \midrule
    \ding{55} & \ding{55} & 21.65 & 3.27 & 21.02 & 17.79 & 12.90 \\
    \ding{51} & \ding{55} & \textbf{27.62} & \textbf{7.02} & 24.03 & \textbf{23.51} & 30.77 \\
    \ding{55} & \ding{51} & 27.04 & 6.41 & 24.15 & \underline{23.88} & \textbf{31.91} \\
    \ding{51} & \ding{51} & \underline{27.49} & \underline{6.96} & \textbf{24.50} & 23.33 & \underline{30.81} \\
    \midrule
    \multicolumn{7}{c}{Trained on MatchTime \& SoccerReplay-1988} \\
    \midrule
    \ding{55} & \ding{55} & 24.17 & 4.09 & 20.51 & 20.70 & 15.70 \\
    \ding{51} & \ding{55} & \underline{28.98} & \textbf{8.39} & 24.45 & \underline{25.35} & \textbf{45.85} \\
    \ding{55} & \ding{51} & 27.54 & 7.76 & \underline{24.50} & 24.70 & 42.79 \\
    \ding{51} & \ding{51} & \textbf{29.21} & \underline{8.22} & \textbf{25.25} & \textbf{25.54} & \underline{43.18} \\
    \bottomrule
    \end{tabular}
    \end{adjustbox}
    \vspace{-3pt}
    \caption{
    \textbf{Ablations on Commentary Generation.} 
    We investigate the impact of different training strategies and datasets on MatchVision using the SoccerReplay-test benchmark.
    `V' and `L' denote the visual encoder and the LLM decoder, respectively.
    }
    \vspace{-12pt}
    \label{tab:ablation_commentary_generation}
\end{table}

\subsection{Qualitative Comparisons}
\label{sec:qualitative_comparison}
As depicted in Figure~\ref{fig:qualitative}, we present qualitative results of MatchVision on the SoccerReplay-test benchmark, comparing models pretrained with and without SoccerReplay-1988.
For event classification, incorporating our data improves accuracy, and even in misclassified cases, the results remain contextually relevant.
For commentary generation, hybrid training on SoccerReplay-1988 enables MatchVision to produce richer, more detailed textual commentary, reflecting a deeper understanding of soccer dynamics.
More qualitative results are available in the \textbf{Appendix}.
\section{Conclusion}
\label{sec:conclusion}
In this paper, we establish a unified, scalable multi-modal framework for soccer understanding. 
Specifically, we introduce \textbf{SoccerReplay-1988}, the largest and most comprehensive soccer video dataset to date, annotated by an automated curation pipeline. 
This provides a solid foundation for developing soccer understanding models and serves as a more challenging benchmark.
Built upon this, we develop \textbf{MatchVision}, an advanced soccer-specific visual encoder, which effectively leverages spatiotemporal information within soccer videos and can be applied to various tasks such as event classification and commentary generation. 
Extensive experiments demonstrate the superiority of our model, with MatchVision achieving state-of-the-art performance on both existing benchmarks and our newly established one. 
We believe this work will set a viable, universal paradigm for future research in sports understanding.
\section*{Acknowledgments}
This work is funded by National Key R\&D Program of China (No.2022ZD0161400).

{
    \small
    \bibliographystyle{ieeenat_fullname}
    \bibliography{main}
}



\onecolumn
{
    \centering
    \Large
    \textbf{Towards Universal Soccer Video Understanding}\\
    \vspace{0.5em} Appendix \\
    \vspace{1.0em}
}
\appendix
{
  \hypersetup{linkcolor=black}
  \tableofcontents
}
\clearpage

\section{SoccerReplay-1988 Dataset Details}
In this section, we provide additional details of our \textbf{SoccerReplay-1988} dataset. 
Specifically, Sec.~\ref{subsec:format-supp} elaborates on the structure and format of the dataset; 
Sec.~\ref{subsec:statistics-supp} presents statistical information and analyses of the dataset;
and Sec.~\ref{subsec:event_summarization} describes the methodology to automatically generate event labels within the dataset.

\subsection{Dataset Format}
\label{subsec:format-supp}

The SoccerReplay-1988 dataset consists of match videos, descriptions of events, and related game information of 1988 soccer matches. 
Each match includes two \texttt{mkv} video files~(1-half and 2-half), covering the match from the initial kick-off to the final whistle.
Additionally, a \texttt{json} file is accompanied by encapsulating detailed information, including event descriptions and comprehensive match backgrounds, structured as follows:

\vspace{2pt}
\noindent \textbf{Match Information} provides background details of the match, including competing teams, final results, and match contexts, such as start time, team formations, and venue details, as illustrated below:

{
\scriptsize
\begin{verbatim}
{
    "timestamp": "2022-08-07 21:00:00",                                # Match start time
    "score": "1 - 2",                                                  # Final score
    "home_team": "Manchester Utd",                                     # Home team name
    "away_team": "Brighton",                                           # Away team name
    "home_formation": "4 - 3 - 3",                                     # Home team formation
    "away_formation": "3 - 4 - 2 - 1",                                 # Away team formation
    "venue": "Old Trafford (Manchester)",                              # Venue and city
    "capacity": "75 635",                                              # Stadium capacity
    "attendance": "73 711",                                            # Number of attendees
}
\end{verbatim}
}
\noindent \textbf{Referee Information} includes details about the primary referee officiating the match, which is formatted as follows:

{
\scriptsize
\begin{verbatim}
{
    "country": "Eng",                                                  # Referee's nationality
    "name": "Paul Tierney"                                             # Referee's name
}
\end{verbatim}
}
\noindent \textbf{Player Information} contains details about various types of individuals involved in the match, including starting players, substitutes, absent players, and coaches. 
All these types are stored in a unified list, with the following format:

{
\scriptsize
\begin{verbatim}
{
    "players_name": "Caicedo M.",                                      # Player's abbreviated name
    "players_number": "25",                                            # Jersey number
    "Full Name": "Moises Caicedo",                                     # Player's full name
    "players_rating": 7.6,                                             # Post-match rating
    "Country": "Ecuador",                                              # Player's nationality
    "Role": "Midfielder",                                              # On-field role
    "Age and Birthdate": "22, (02.11.2001)",                           # Age and birth date
    "Market Value": "€89.4m"                                           # Player's market value
}
\end{verbatim}
}
\noindent \textbf{Event Descriptions} is a list that records all key events during the match, including their types and detailed commentary. 
A typical example of an event entry is shown below:
{
\scriptsize
\begin{verbatim}
{
    "half": 1,                                                         # Match half (1 or 2)
    "time_stamp": "00:16",                                             # Timestamp within the half
    "comments_type": "shot off target",                                # Event type
    "comments_text": "A mistake by Leandro Trossard (Brighton)...",    # Commentary text 
    "comments_text_anonymized": "A mistake by [PLAYER]([TEAM])..."     # Commentary after anonymization
}
\end{verbatim}
}

\subsection{Additional Dataset Statistics}
\label{subsec:statistics-supp}

\noindent
\begin{minipage}{0.48\textwidth}
\centering
\small
\begin{tabular}{lc}
\toprule
\textbf{League}    & \textbf{\# Match} \\ 
\midrule
Italy Serie-a            & 367 \\
England Premier League   & 552 z\\
UEFA Champions League    & 469 \\
France Ligue 1           & 123 \\
Spain LaLiga             & 235 \\
Germany Bundesliga       & 242 \\
\bottomrule
\end{tabular}
\vspace{7pt}
\captionof{table}{\textbf{League-wise Match Statistics.}}
\label{tab:league-wise}
\end{minipage}
\begin{minipage}{0.48\textwidth}
\centering
\small
\begin{tabular}{lc}
\toprule
\textbf{Season}    & \textbf{\# Match} \\ 
\midrule
\textit{2017-2018}          & 172 \\
\textit{2018-2019}          & 325 \\
\textit{2019-2020}          & 300 \\
\textit{2020-2021}          & 323 \\
\textit{2021-2022}          & 330 \\
\textit{2022-2023}          & 416 \\
\textit{2023-2024}          & 122 \\
\bottomrule
\end{tabular}
\vspace{-2pt}
\captionof{table}{\textbf{Season-wise Match Statistics.}}
\label{tab:year-wise}
\end{minipage}

\noindent
To provide a comprehensive analysis of \textbf{SoccerReplay-1988}, we present statistics and visualizations in the following tables and figures. 
Concretely, Table~\ref{tab:league-wise} and~\ref{tab:year-wise} illustrate the distribution of the 1988 matches across different leagues and seasons. 
Figure~\ref{fig:visualization}~(a) compares SoccerReplay-1988 with other soccer datasets~\cite{giancola2018soccernet1, deliege2021soccernet2, rao2024matchtimeautomaticsoccergame, goal, jiang2020soccerdb, yu2018fine, suglia2022going}, highlighting its unique scale.
Figure~\ref{fig:visualization}~(b) depicts the distribution of event labels for the 24 newly defined categories. 
Finally, Figures~\ref{fig:visualization}~(c),~(d),~(e), and~(f) present detailed analyses of commentary data, including frequency distributions, timestamps, and word counts.

\begin{figure}[h!]
\centering
\includegraphics[width=0.97\textwidth]{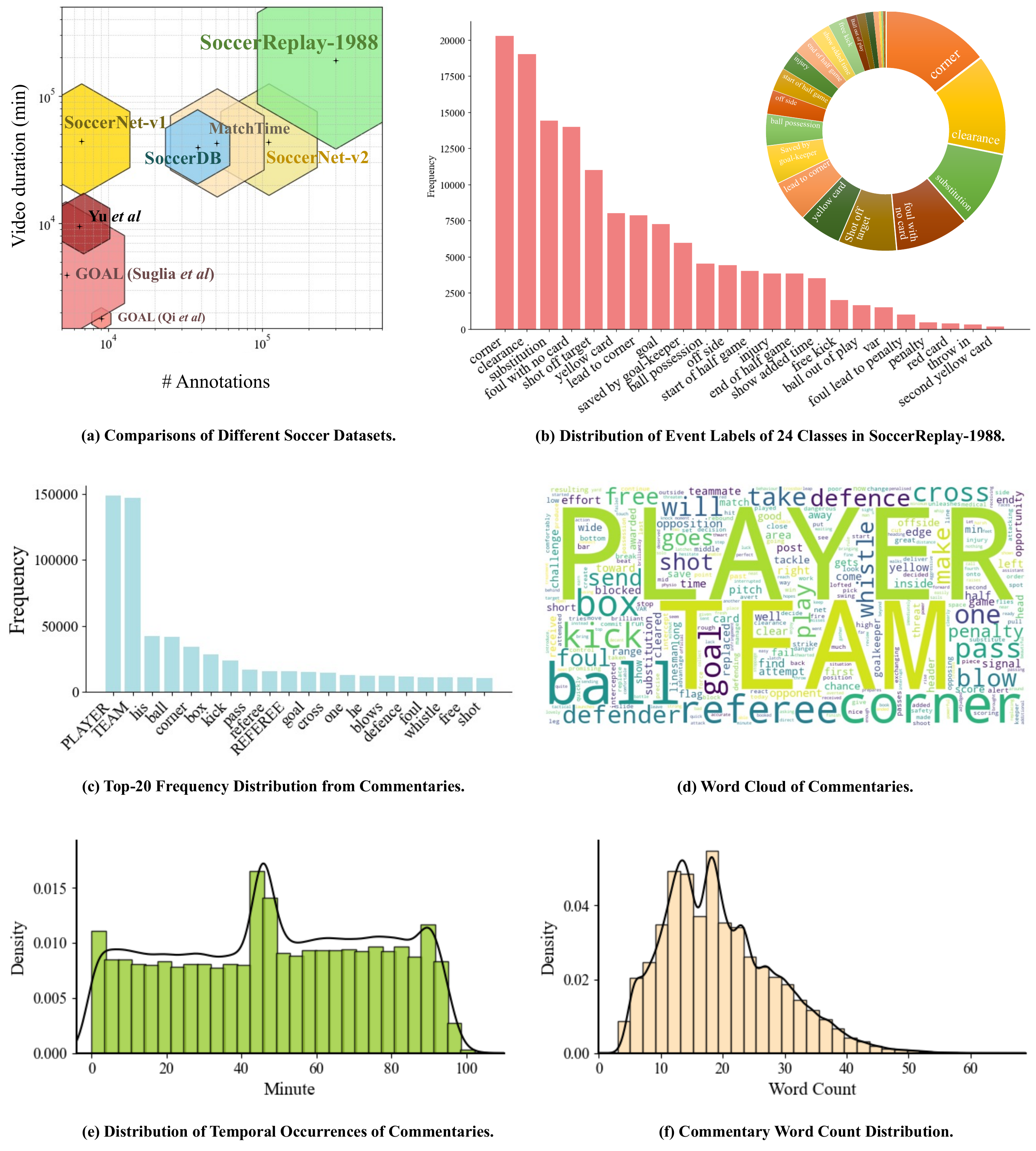} 
\caption{\textbf{Comprehensive Visualizations of SoccerReplay-1988 Dataset.}}
\label{fig:visualization}
\end{figure}

\subsection{Event Summarization}
\label{subsec:event_summarization}
Our dataset expands the original 17 event categories in SoccerNet~\cite{deliege2021soccernet2} to 24 types. 
For MatchTime~\cite{rao2024matchtimeautomaticsoccergame} and SoccerReplay-1988, LLaMA-3 (70B)~\cite{llama3} is employed to analyze commentaries and generate corresponding event labels. 
This is guided by a carefully refined prompt, iteratively improved through manual checks and enriched with comprehensive definitions and examples.
Notably, text commentaries unrelated to visual content~({\em e.g.,} {\em `possession ratio is 55:45'}) are categorized as {\em `statistics and summary'}, and excluded from model training and testing.
Finally, a random sample of 2\% of the data yields 98\% manual verification accuracy, confirming the high quality of automated labeling. 
The entire prompt is presented below:
\vspace{6pt}

{
\scriptsize
\begin{tcolorbox}[mybox]
\texttt{<|begin\_of\_text|>}
\\ 
\texttt{<|start\_header\_id|>system<|end\_header\_id|>}

\ \

You are an expert in soccer, you have a very important task to summarize a soccer commentary into certain types of events. The accuracy of your classification is the most emergency thing. I will give you a commentary sentence. You need to select one type of event that can best describe this event from the following 24 types: \textit{'corner', 'goal', 'injury', 'own goal', 'penalty', 'penalty missed', 'red card', 'second yellow card', 'substitution', 'start of game(half)', 'end of game(half)', 'yellow card', 'throw in', 'free kick', 'saved by goal-keeper', 'shot off target', 'clearance', 'lead to corner', 'off-side', 'var', 'foul (no card)', 'statistics and summary', 'ball possession', 'ball out of play'}.

\ \

Here are some rules you have to obey when summarizing types, you should consider it strictly following these steps: 

\ \ 

\begin{enumerate}
    \item Firstly, you need to find if there is any evidence of foul in commentary, if yes, it can only be \textit{'foul (no card)'}, \textit{'yellow card'}, \textit{'red card'} or \textit{'second yellow card'} according to the situation, even though it introduces the result \textit{'free kick'} or \textit{'penalty'}. For example: 'Per Mertesacker (Arsenal) commits a rough foul. Michael Dean stops the game and makes a call. That's a free kick to Manchester Utd.' can ONLY be \textit{'foul (no card)'} since there is a foul in commentary, even though the result is \textit{'free kick'}.
    \vspace{2pt}
    \item Secondly, only if the word \textit{'corner'} is in the commentary, you need to select it from \textit{'lead to corner'}. \textit{'lead to corner'} means the process of how the \textit{'corner'} occurs, which is before the \textit{'corner'} kick. For type \textit{'lead to corner'}, there will always be words like 'award a \textit{corner}', 'will have a \textit{corner}', 'point at \textit{corner} flag' and so on. For example: 'Victor Wanyama (Southampton) goes on a solo run, but he fails to create a chance as an opposition player blocks him. The referee signals a \textit{corner} kick to Southampton.' is \textit{'lead to corner'}.
    \vspace{2pt}
    \item Thirdly is the most easy-confused part, you need to be cautious: only if the word \textit{'free-kick'}/\textit{'free kick'} is in the commentary will it be a \textit{'free kick'}. According to the first rule, if there is foul in the sentence, it cannot be \textit{'free kick'}. \textit{'free kick'} can only be selected when \textit{'free-kick'}/\textit{'free kick'} occurs in commentary and is describing the process of a \textit{'free kick'} attack. For example: 'Olivier Giroud (Arsenal) gets on the ball and beats an opponent, but his run is stopped by the referee Michael Dean who sees an offensive foul. It's a \textit{free kick} to Burnley, but they probably won't attempt a direct shot on goal from here.' is \textit{'foul (no card)'}; 'Ander Herrera (Manchester United) makes a slide tackle, but referee Michael Dean blows for a foul. \textit{Free kick}. Arsenal will probably just try to cross the ball in from here.' is \textit{'foul (no card)'}; 'Marcos Rojo (Manchester United) connects with the \textit{free kick} and produces a header goalwards which is well blocked. The goalkeeper doesn't have to worry about that one.' is \textit{'free kick'}.
    \vspace{2pt}
    \item Similarly, \textit{'penalty'} and \textit{'penalty missed'} only describe things that happen during a \textit{'penalty'} kick. If it is introducing the reason that leads to a \textit{'penalty'}, you should return the type describing the reason, like \textit{'foul (no card)'}, \textit{'yellow card'}, and so on.
    \vspace{2pt}
    \item The type \textit{'statistics and summary'} includes all the commentaries that are not introducing visually evidential events, but those statistics or overviews of the game. These sentences won’t concentrate on certain events, but on the overall game.
    \vspace{2pt}
    \item \textit{'ball possession'} represents those commentaries that describe any of the teams controlling the \textit{'ball possession'}.
    \vspace{2pt}
    \item You need to be sensitive about the type \textit{'shot off target'}; if there is an event of a shot happening in the commentary, it is a shot. If it's not a \textit{'goal'}, didn't make a score, and was not saved by the goalkeeper, it would probably be a \textit{'shot off target'}. Normally there will be keywords like 'wide of the right post', 'over the crossbar', 'crashes against the crossbar' and so on. You have to judge it sensitively about the situation after the shot.
    \vspace{2pt}
    \item An important type after a shot: \textit{'saved by goal-keeper'} describes that the shot is saved by the goalkeeper; there would be words like 'blocked', 'saved', and so on. Especially when \textit{'goal-keeper'}/\textit{'goal keeper'} occurs in the commentary sentence!!! it will probably be \textit{'saved by goal-keeper'}. You need to find it carefully!!!
    \vspace{2pt}
    \item If a player lofts or swings a pass to a penalty area/dangerous area, they might be \textit{'shot off target'}, \textit{'clearance'}, \textit{'saved by goal-keeper'}, and so on. It should NOT be identified as \textit{'corner'} or \textit{'free kick'} if there is no obvious evidence in commentary! For example: 'Tomas Rosicky (Arsenal) fails to find any of his teammates inside the box as his pass is blocked.' should be \textit{'clearance'} rather than \textit{'corner'} or \textit{'free kick'}.
    \vspace{2pt}
    \item \textit{'clearance'} means those good performances in defense; they stop the offense of opponents. If such a successful defense happens in the commentary, it can only be \textit{'clearance'}. In these commentaries, there are always some words like 'opponent's defense', 'intercepts the ball', 'clear the ball', and so on.
    \vspace{2pt}
    \item \textit{'offside'} is an obvious event; there are always the words 'flag', 'linesman', 'too fast to defense' in the commentary since \textit{'offside'} is the player running forward the defense line, and the linesman will raise the flag.
    \vspace{2pt}
    \item \textit{'ball out of play'} means any player kicks the ball out of boundary lines. These commentary sentences will mostly end up with throw-ins or goal kicks.
    \vspace{2pt}
    \item \textit{'throw-in'} means exactly the process of \textit{'throw-in'} balls.
    \vspace{2pt}
    \item Most \textit{'goals'} are normal \textit{'goals'}. If you see a scoring event, you can only identify the score as \textit{'own goal'} when there is obvious evidence.
\end{enumerate}

\ \

\texttt{<|eot\_id|>}  
\\ 
\texttt{<|start\_header\_id|>user<|end\_header\_id|>}

\ \

With the classification rules, you should tell me the type of a commentary from above candidate options: \textit{'corner', 'goal', 'injury', 'own goal', 'penalty', 'penalty missed', 'red card', 'second yellow card', 'substitution', 'start of game(half)', 'end of game(half)', 'yellow card', 'throw in', 'free kick', 'saved by goal-keeper', 'shot off target', 'clearance', 'lead to corner', 'off-side', 'var', 'foul (no card)', 'statistics and summary', 'ball possession', 'ball out of play'}. The commentary sentence you need to define type is: 

\ \ 

\quad\quad [COMMENTARY TEXT HERE (before anonymization)] 

\ \ 

You need to carefully consider the rules in order and make your final decision. Now, you must return me the name of its type from candidate options (in lower case, only return the name of type, answer it right away after my prompt without any other words). 

\ \

\texttt{<|eot\_id|>}
\end{tcolorbox}
}

\newpage

\section{SoccerNet-pro Dataset Details}
As discussed in the main text, 
alongside the SoccerReplay-1988 dataset, we also incorporate two existing datasets, SoccerNet-v2~\cite{deliege2021soccernet2} and MatchTime~\cite{rao2024matchtimeautomaticsoccergame} to enrich the training data. 
These datasets undergo the following preprocessing strategies and are then unified into the SoccerNet-pro dataset, ensuring format consistency with SoccerReplay-1988.

\subsection{SoccerNet-v2}
\label{subsec:soccernet-v2_labels-supp}
The SoccerNet-v2~\cite{deliege2021soccernet2} dataset comprises over 110k event labels across 500 matches, categorized into 17 distinct classes. 
Based on soccer rules and specific domain knowledge, these labels are systematically reclassified into 24 categories with our proposed automated data curation pipeline, as detailed in Table \ref{tab:process_strategy}.

\begin{table}[ht]
\centering
\footnotesize
\renewcommand{\arraystretch}{1} 
\setlength{\tabcolsep}{8pt} 
\begin{tabular}{l@{\hskip 10mm}l@{\hskip 10mm}l}
\toprule 
\textbf{Original Label}        & \textbf{Processed Label}      & \textbf{Reference}                                                                 \\ \midrule 
\multirow{2}{*}{Penalty}       & Penalty                       & Scored penalties are categorized as ``Penalty.''                                   \\
                               & Penalty Missed                & Missed penalties are categorized as ``Penalty Missed.''                            \\ \midrule
Kick-off                      & Start of Game (Half)          & Matches the start of a half after goals.                                           \\ \midrule
Shots off target              & Shot Off Target               & No change.                                                          \\ \midrule
Throw-in                      & Throw In                      & No change.                                                          \\ \midrule
Ball out of play              & Ball Out of Play              & No change.                                                          \\ \midrule
Foul                          & Foul (No Card)                & Refers to fouls without cards for only.                                            \\ \midrule
Yellow card                   & Yellow Card                   & No change.                                                          \\ \midrule
Yellow$\rightarrow$red card & Second Yellow Card            & No change.                                                          \\ \midrule
Red card                      & Red Card                      & No change.                                                          \\ \midrule
Direct free-kick              & \multirow{2}{*}{Free Kick}    & \multirow{2}{*}{Both direct and indirect free kicks are grouped.}                  \\ 
Indirect free-kick            &                               &                                                                                     \\ \midrule
Substitution                  & Substitution                  & No change.                                                          \\ \midrule
Goal                          & Goal                          & No change.                                                          \\ \midrule
Clearance                     & Clearance                     & No change.                                                          \\ \midrule
Offside                       & Off-Side                      & No change.                                                          \\ \midrule
Corner                        & Corner                        & No change.                                                          \\ \bottomrule 
\end{tabular}
\caption{
\textbf{Processing Strategy for SoccerNet-pro.} 
The \textbf{Reference} column details the specific processing applied to the original labels.
}
\label{tab:process_strategy}
\end{table}
\vspace{-10pt}

\subsection{MatchTime}
The MatchTime dataset~\cite{rao2024matchtimeautomaticsoccergame}, curated from SoccerNet-Caption~\cite{densecap}, contains a substantial amount of commentary, with only a small portion accompanied by event labels.
To bridge this gap, we apply the prompt-based approach described in Sec.~\ref{subsec:event_summarization} to summarize commentaries into event labels, assigning each commentary a corresponding label.

\vspace{2pt}
\subsection{Data Split Strategy}
As described in the manuscript, 
SoccerReplay-1988 is divided into 1,488 matches for training, 250 for validation, and 250 for testing.
For the processed SoccerNet-pro dataset~(including SoccerNet-v2 and MatchTime), we adhere to the original partitioning strategies and match distributions of its source datasets as detailed in Table~\ref{tab:dataset_splits}.

\begin{table}[ht]
    \centering
    \renewcommand{\arraystretch}{1.3} 
    \begin{tabular}{lcccc}
        \toprule 
        \textbf{Dataset} & \textbf{Train} & \textbf{Valid} & \textbf{Test} & \textbf{Total} \\
        \midrule 
        SoccerNet-v2~\cite{deliege2021soccernet2} & 300  & 100 & 100 & 500 \\
        MatchTime~\cite{rao2024matchtimeautomaticsoccergame} & 373  & 49  & 49  & 471 \\
        \textbf{SoccerReplay-1988} & 1488 & 250 & 250 & 1988 \\
        \bottomrule 
    \end{tabular}
    \caption{\textbf{Dataset Splits for Training, Validation, and Testing.}}
    \vspace{-8pt}
    \label{tab:dataset_splits}
\end{table}

\section{Implementation Details}
In this section, we provide additional implementation details about MatchVision.
Sec.~\ref{subsec:data_preprocessing} presents more information on data preprocessing strategies;
Sec.~\ref{subsec:strategy-supp} elaborates on the evaluation strategies used during model training;
and Sec.~\ref{subsec:hyperparameter_selection} discusses several hyperparameter choices inspired by prior works.

\subsection{Data Preprocessing}
\label{subsec:data_preprocessing}
Our automated data curation pipeline filters out video clips with missing annotations, incorrect cropping, or invalid timestamps.
In all experiments, video frames are resized to $224\times224$ and preprocessed using the image preprocessor of SigLIP~\cite{zhai2023SigLIP}, which normalizes frames to a mean of 0.5 and a standard deviation of 0.5 before serving as inputs.
For overlapping video content between SoccerNet-v2~\cite{deliege2021soccernet2} and MatchTime~\cite{rao2024matchtimeautomaticsoccergame}, we prioritize using event labels from SoccerNet-v2.

\subsection{Validation Strategy during Training}
\label{subsec:strategy-supp}
We select the best-performing checkpoints on the validation set with the evaluation strategies detailed below:

During pretraining:
(i) For \textbf{supervised classification}, we adopt top-1/3/5 event classification accuracy on the validation set to select the best model;
(ii) For \textbf{visual-language contrastive learning}, video-to-text retrieval is performed, with top-1/3/5 accuracy of event classification~(comparing retrieved texts’ event labels to ground truth) as the validation metric.

During downstream tasks training:
(i) In \textbf{event classification} and \textbf{foul recognition}, classification accuracy on the validation set is used as the evaluation metric; 
(ii) For \textbf{commentary generation}, the CIDEr~\cite{vedantam2015cider} score of the model's predictions on the validation set is employed to select the best checkpoint.

\subsection{Hyperparameter Selection}
\label{subsec:hyperparameter_selection}
Here, we provide further explanations about the hyperparameters in our model, inspired by prior works, as detailed below:

\vspace{2pt}
\noindent \textbf{Temporal Window Size.}
We adopt a 30-second temporal window to extract video clips.
This is inspired by MatchTime~\cite{rao2024matchtimeautomaticsoccergame}, which demonstrates that a 30-second window is sufficient to capture adequate visual information for optimal performance, outperforming the 45-second window used in SoccerNet-Caption~\cite{densecap}.

\vspace{2pt}
\noindent \textbf{LoRA Rank.}
For finetuning the commentary generation head, we use LoRA~\cite{hu2021lora} with a rank of 16, following~\cite{rao2024matchtimeautomaticsoccergame}.

\vspace{2pt}
\noindent \textbf{Query Length of Perceiver.}
For the Perceiver~\cite{jaegle2021perceiver} module in the commentary generation head, we utilize a query length of 32 for temporal information aggregation, consistent with the optimal configuration reported in~\cite{rao2024matchtimeautomaticsoccergame}.

\begin{figure}[htp]
\centering
\includegraphics[width=0.8\textwidth]{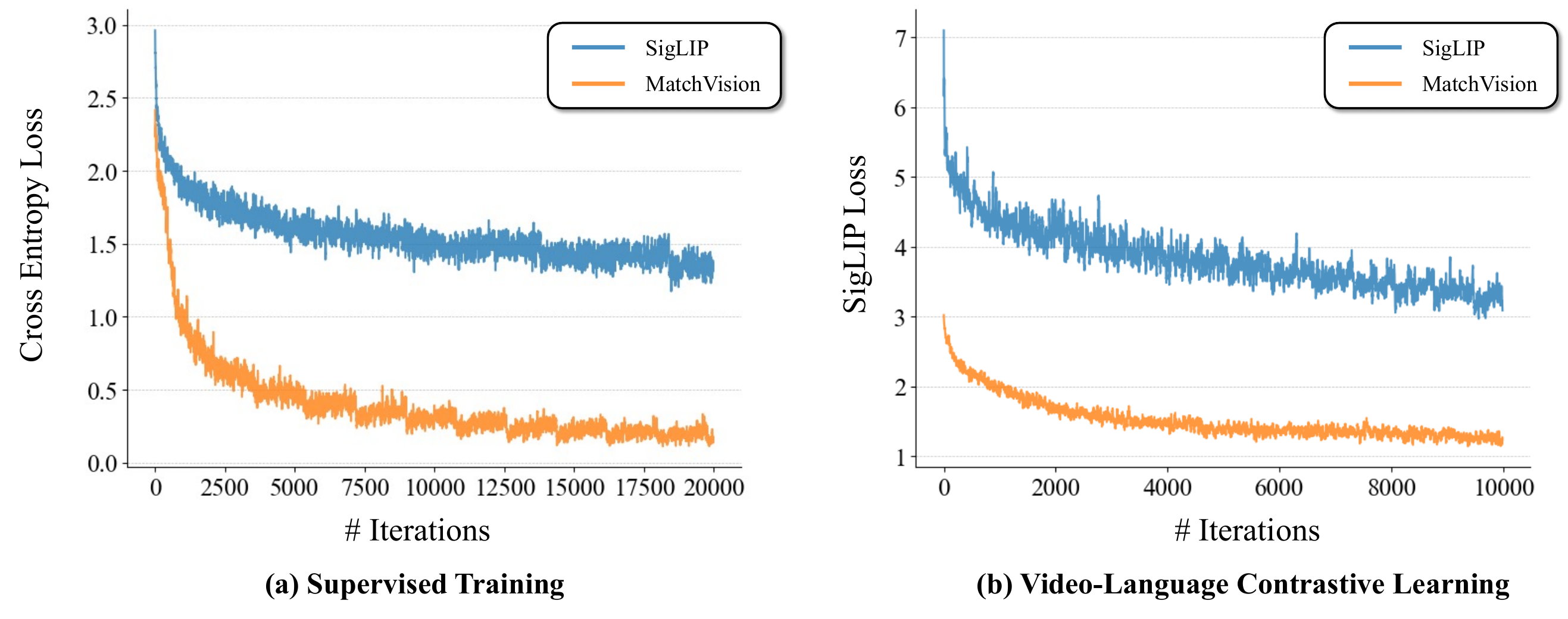}
\vspace{-9pt}
\caption{\textbf{Training Loss Curves of Visual Encoders Pretraining.}}
\vspace{-15pt}
\label{fig:curve}
\end{figure}

\section{Experiments}
In this section, we provide additional details to offer deeper insights into our model and its performance. 
Specifically, Sec.~\ref{subsec:curves-supp} presents training curves to clearly illustrate the training process; 
Sec.~\ref{subsec:quantitative-supp} and Sec.~\ref{subsec:qualitative-supp} showcase more quantitative and qualitative results, respectively, demonstrating the model's capability to effectively understand soccer dynamics.

\subsection{Training Curves}
\label{subsec:curves-supp}
We present the loss curves for visual encoder pretraining in Figure~\ref{fig:curve}. 
Our MatchVision demonstrates significantly better convergence compared to SigLIP~\cite{zhai2023SigLIP} backbone, indicating that it effectively leverages spatiotemporal attention to utilize temporal information, learning representations better suited for highly dynamic soccer videos.

\subsection{More Quantitative Results}
\label{subsec:quantitative-supp}
Here, we compare present comparisons with more advanced methods in the SoccerNet Foul Recognition challenge~\cite{cioppa2024soccernet2024challengesresults}, where MatchVision remains competitive even with \textbf{a frozen visual encoder}.

\begin{table}[htp]
\centering
\small
\setlength{\tabcolsep}{3pt} 
\renewcommand{\arraystretch}{1.1} 
\begin{tabular}{c|cccccccc}
\toprule
\textbf{Top-1 Accuracy} & \textbf{Ours} & Baseline & zyz & PD\_PS\_GSN & Redsox & xiao\_he\_shang \\ 
\midrule
\textbf{Foul} & 0.44 & 0.36 & 0.58 & 0.44 & \textbf{0.60} & 0.46 \\ 
\textbf{Severity} & \textbf{0.58} & 0.54 & 0.58 & 0.50 & 0.05 & 0.47 \\
\bottomrule
\end{tabular}
\vspace{-0.15cm}
\caption{\textbf{More Quantitative Results on Multi-view Foul Recognition.}}
\vspace{-0.35cm}
\end{table}

\subsection{More Qualitative Results}
\label{subsec:qualitative-supp}
More qualitative visualizations of commentary generation across various events on the field are depicted in Figure~\ref{fig:more_qualitative_results_1},~\ref{fig:more_qualitative_results_2}, and~\ref{fig:more_qualitative_results_3}.

\section{Limitations \& Future Work}
Although MatchVision explores establishing a soccer-specific visual encoder, it is not without its limitations:
(i) Currently, MatchVision is adapted to event classification, commentary generation, and foul recognition tasks. 
In the future, we plan to further extend it to more challenging tasks such as player tracking and dense video captioning, aiming to develop a more comprehensive foundation model for soccer analysis.
(ii) Given computational and annotation constraints, SoccerReplay-1988 primarily focuses on European league soccer data. 
We aim to leverage our scalable automated annotation pipeline to further expand the dataset, encompassing a more comprehensive range of soccer data.
(iii) Following prior works, our commentary generation remains anonymized.
This is left for future work, where we aim to fully leverage contextual information available in our SoccerReplay-1988 dataset to enable more vivid, accurate, and context-aware commentary generation.

\begin{figure}[hb]
  \centering
  \includegraphics[width=0.88\textwidth]{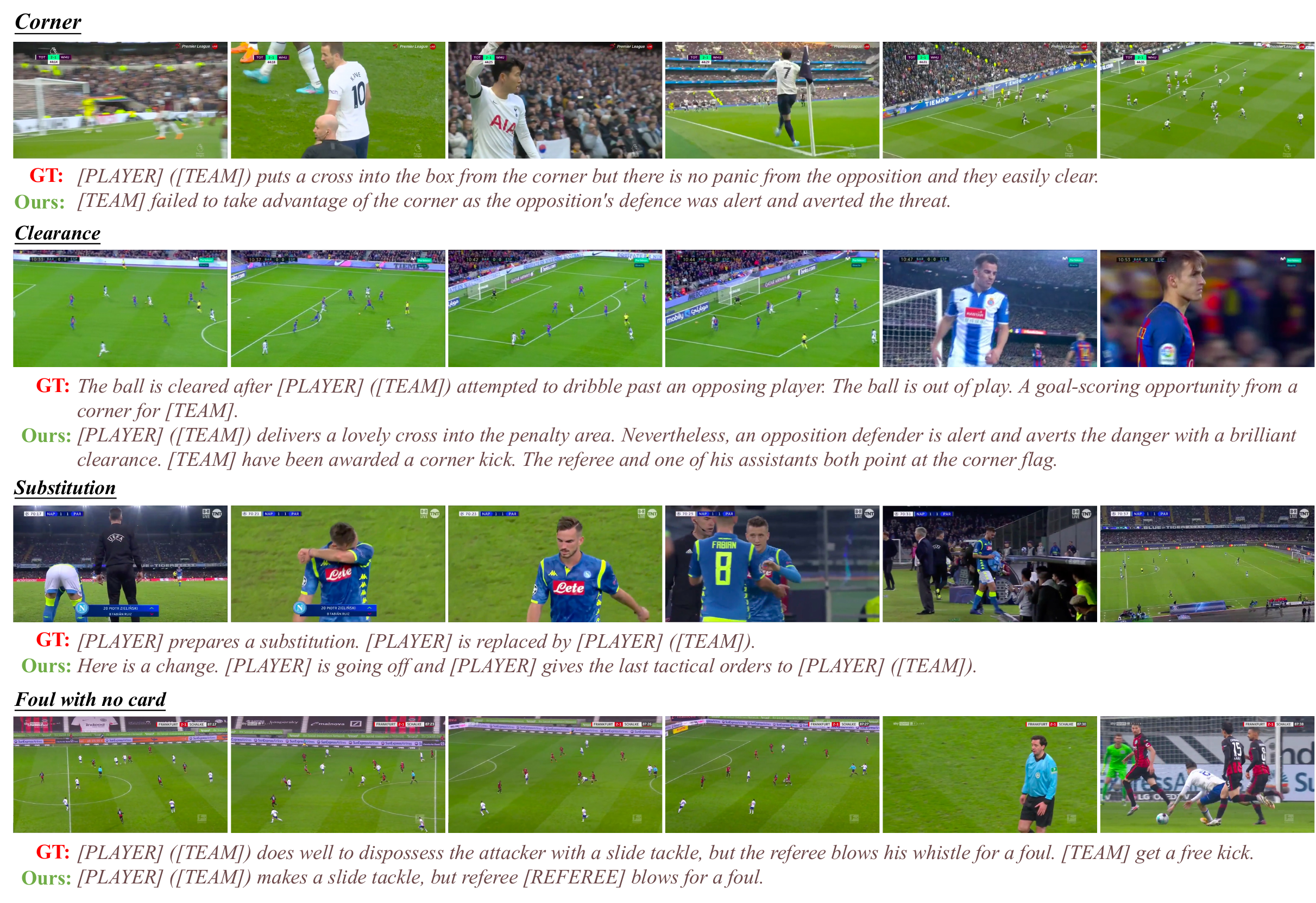} \\
  \vspace{-4pt}
  \caption{
  \textbf{More Qualitative Results of Commentary Generation.}
  }
    \vspace{-4pt}
 \label{fig:more_qualitative_results_1}
\end{figure}

\begin{figure}[p]
  \centering
  \includegraphics[width=0.88\textwidth]{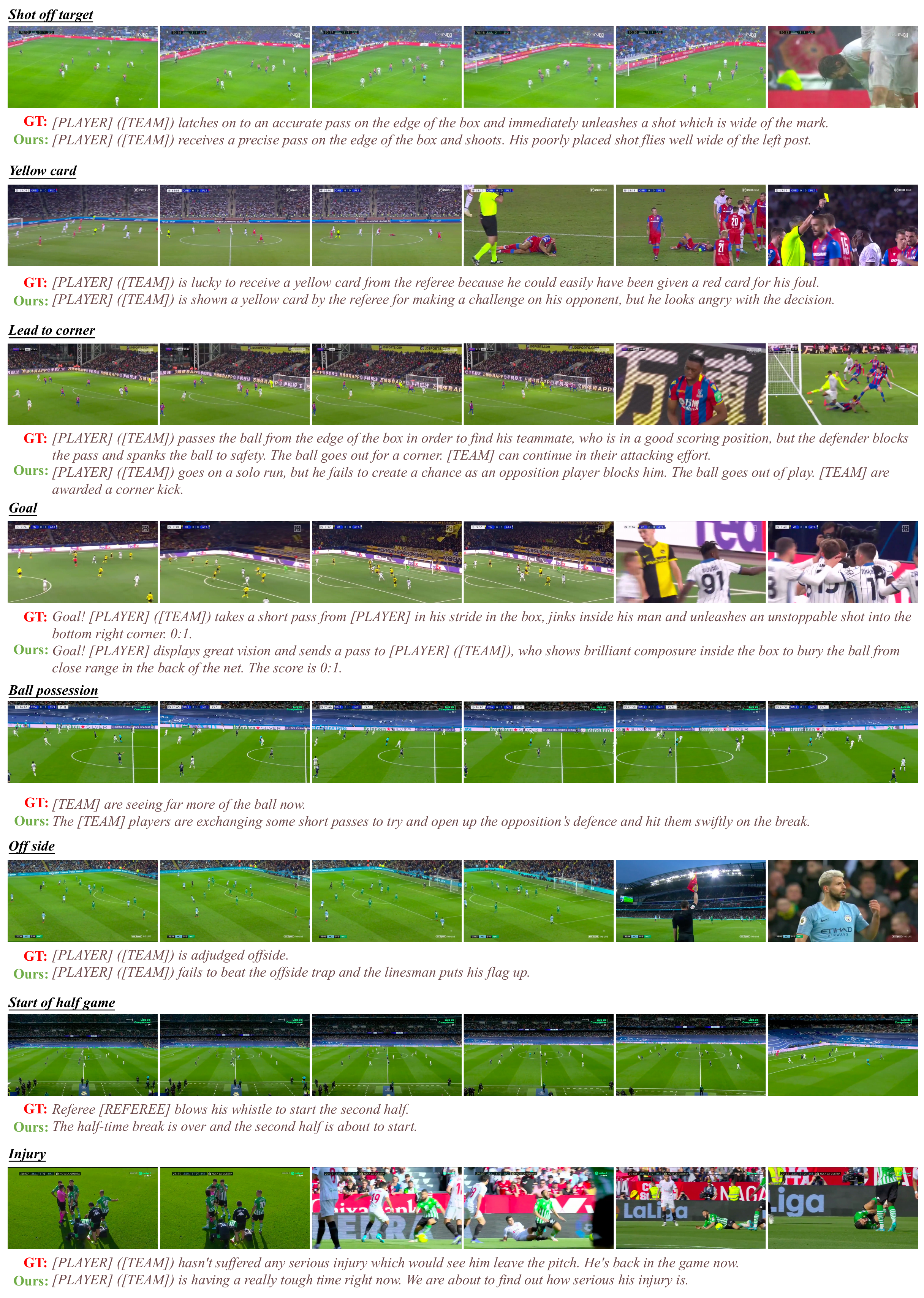} \\
  \vspace{-2pt}
  \caption{
  \textbf{More Qualitative Results of Commentary Generation.}
  }
    \vspace{-4pt}
 \label{fig:more_qualitative_results_2}
\end{figure}

\begin{figure}[p]
  \centering
  \includegraphics[width=0.88\textwidth]{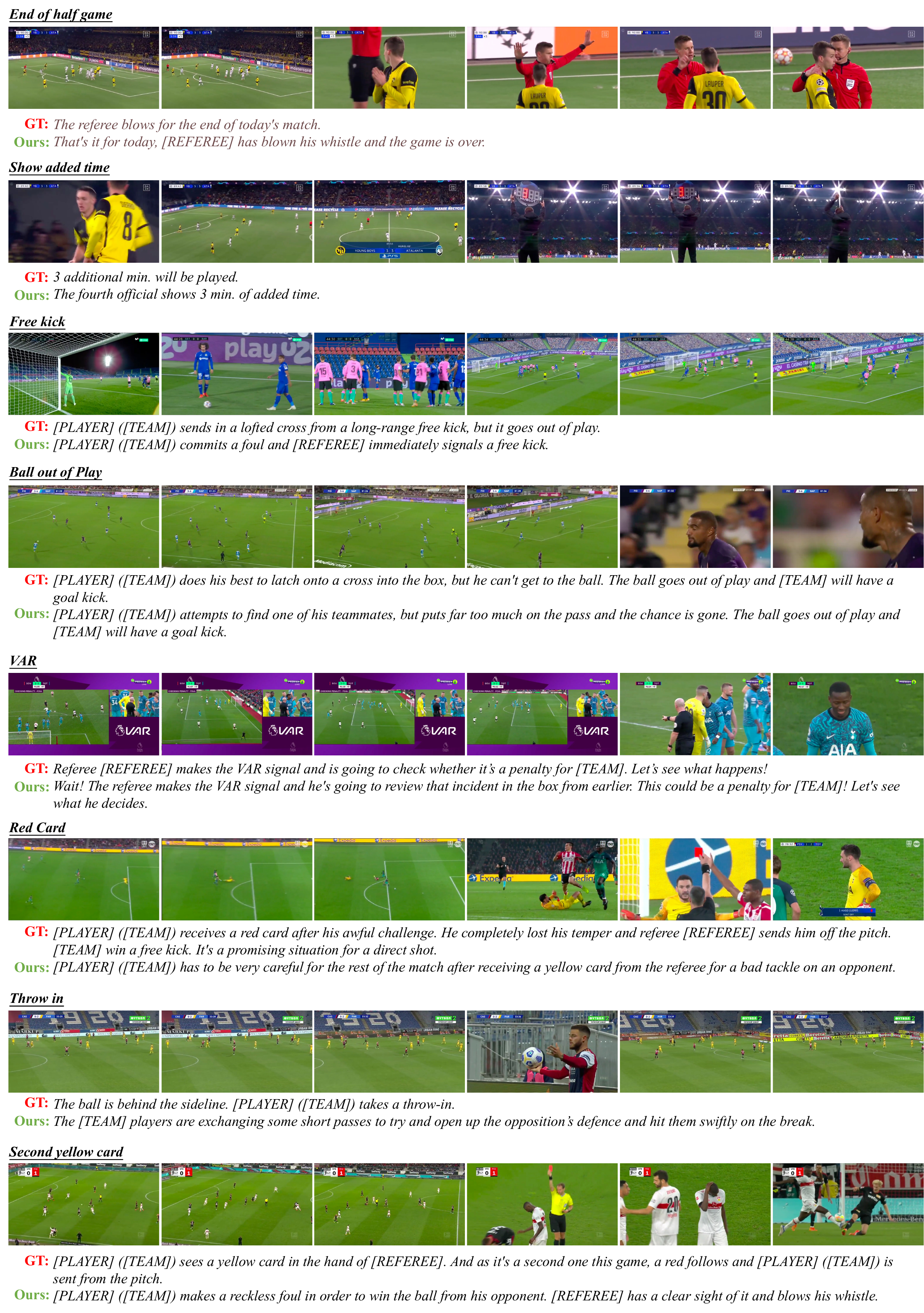} \\
  \vspace{-2pt}
  \caption{
  \textbf{More Qualitative Results of Commentary Generation.}
  }
    \vspace{-4pt}
 \label{fig:more_qualitative_results_3}
\end{figure}

\clearpage

\end{document}